\documentclass[a4paper,fleqn]{cas-dc}

\usepackage[numbers]{natbib}
\usepackage{subfigure}
\usepackage{lscape}

\def\tsc#1{\csdef{#1}{\textsc{\lowercase{#1}}\xspace}}
\tsc{WGM}
\tsc{QE}
\tsc{EP}
\tsc{PMS}
\tsc{BEC}
\tsc{DE}

\begin{document}
\let\WriteBookmarks\relax
\def\floatpagepagefraction{1}
\def\textpagefraction{.001}

\shorttitle{}

\shortauthors{Zhengbin Zhang}

\title [mode = title]{Cross-CBAM: A Lightweight network for Scene Segmentation}                      
\tnotemark[]




%
\author[1]{Zhengbin\ Zhang}[type=editor,
                        orcid=0000-0002-4747-1147]

\cormark[0]


\ead{y80210135@main.ecust.edu.cn}


\credit{Conceptualization of this study, Methodology, Software, Writing-original draft}

\affiliation[1]{organization={East China University Of Science And Technology},
    addressline={No.130, MeiLong Road, XuHui District}, 
    city={Shanghai},
    postcode={200237}, 
    country={China}}

\author[1]{Zhenhao\ Xu}[]
\credit{Review \& editing, Investigation, Project administration}

\author[1]{Xingsheng\ Gu}[]
\credit{Supervision, Review \& editing, Supervision}

\author[2]{Juan\ Xiong}[]
\credit{Hardware support, Review \& editing}
\affiliation[2]{organization={University of Shanghai for Science and Technology},
    addressline={No.516, JunGon Road, YangPu District}, 
    city={Shanghai},
    postcode={200093}, 
    country={China}}








\begin{abstract}
Scene parsing is a great challenge for real-time semantic segmentation. Although traditional semantic segmentation networks have made remarkable leap-forwards in semantic accuracy, the performance of inference speed is unsatisfactory. Meanwhile, this progress is achieved with fairly large networks and powerful computational resources. However, it is difficult to run extremely large models on edge computing devices with limited computing power, which poses a huge challenge to the real-time semantic segmentation tasks. In this paper, we present the Cross-CBAM network, a novel lightweight network for real-time semantic segmentation. Specifically, a Squeeze-and-Excitation Atrous Spatial Pyramid Pooling Module(SE-ASPP) is proposed to get variable field-of-view and multiscale information. And we propose a Cross Convolutional Block Attention Module(CCBAM), in which a cross-multiply operation is employed in the CCBAM module to make high-level semantic information guide low-level detail information. Different from previous work, these works use attention to focus on the desired information in the backbone. CCBAM uses cross-attention for feature fusion in the FPN structure. Extensive experiments on the Cityscapes dataset and Camvid dataset demonstrate the effectiveness of the proposed Cross-CBAM model by achieving a promising trade-off between segmentation accuracy and inference speed. On the Cityscapes test set, we achieve $73.4\%$ mIoU with a speed of $240.9$FPS and $77.2\%$ mIoU with a speed of $88.6$FPS on NVIDIA GTX 1080Ti.
\end{abstract}




\maketitle

\section{Introduction}
Semantic segmentation is one of the main tasks of computer vision, where the goal is to classify all pixels in an image with a dense label classification. Advances in deep learning have greatly boosted the accuracy of semantic segmentation in several benchmarks, e.g., Pascal Voc\cite{Everingham2010}, ADE 20k\cite{Zhou2019}, Microsoft COCO\cite{Lin2014}, and Cityscapes\cite{Cordts2016}. Due to its excellent performance, it has been widely used in autonomous driving, medical image segmentation, machine vision and other fields. However, these applications place higher requirements for mobile deployment.

To meet the demands of these fields, a lot of researchers have proposed real-time semantic segmentation models with low parameter quantity and high inference speed. These models reduce inference time from two perspectives:1) try to restrict the input image size \cite{wu2017real}, or prune redundant channels in the networks \cite{badrinarayanan2017segnet}\cite{paszke2016enet}\cite{zhao2018icnet} to reduce computation complexity and boost the inference speed. Fewer channels and smaller input resolution of an image seem to be effective (or work), however, it can easily lose spatial details around boundaries, corners, and small objects, leading to the loss of  metric accuracy. 2) try to choose lightweight backbones \cite{chollet2017xception,howard2019searching} to reduce parameters and Flops. Most of these lightweight networks take advantage of the ImageNet1k pre-trained weights for faster training and higher accuracy. But, this approach poses two serious problems. First, ImageNet1k has one thousand classes, to classify each image, the backbone used to train ImageNet1k typically contains thousands of channels in the last few convolutional layers, resulting in redundant channels. The second point is that the input size of ImageNet1k is smaller than the segmentation, resulting in the lack of sufficient receptive field and detail information in the ImageNet1k network.

To tackle the above problems, BiSeNet\cite{yu2018bisenet} and BiSeNetV2 \cite{yu2021bisenet} propose a multi-path structure to extract low-level details and high-level semantics. U-Net\cite{ronneberger2015u} adopts the U-shape framework to gradually reduce channels and increase spatial resolution. Therefore, it can compensate some spatial details lost due to downsampling. DDRNet\cite{hong2021deep} proposes a dual-resolution network with two parallel deepth branches with different resolution. But adding another path increases model complexity and inference time. At the same time, the auxiliary path always lacks low-level information guidance.

\begin{figure*}[htbp]
    \includegraphics[width=1\textwidth,height=0.7\textwidth]{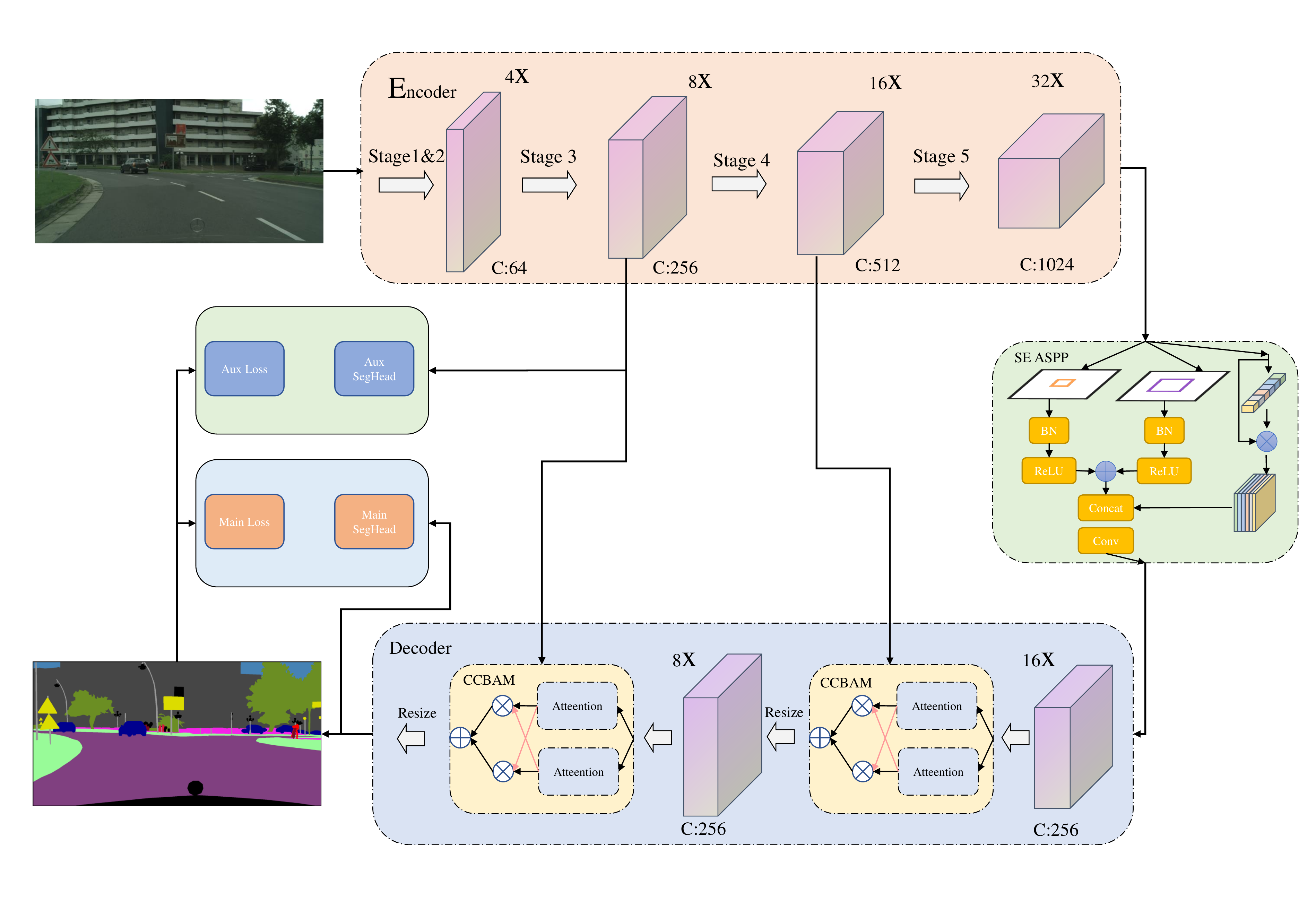}
    \caption{\textbf{Overview of the proposed architecture.} \textit{Encoder} denotes the STDC network, which consists of five stages, \textit{Decoder} denotes the proposed method, \textit{CCBAM} and \textit{SE-ASPP} denote the innovative structures, respectively. \textit{Conv\&BN} are used for channel transformation, and \textit{Resize} means bilinear interpolation upsampling. \textit{C} represents the number of channels, and \textit{4X} refers to the multiple of downsampling the original image.}
    \label{fig1_overview_of_crosscbam}
\end{figure*}

In this paper, we propose a novel hand-craft network with the aim of improving higher segmentation accuracy, explainable structure, and competitive performance to that of existing methods. The network is named Cross-CBAM. As illustrated in Figure \ref{fig1_overview_of_crosscbam}, the Cross-CBAM network adopts the encode-decode architecture and consists of two novel modules: Squeeze-and-excitation Atrous Spatial Pyramid Pooling Module(SE-ASPP) and Cross Convolutional Block Attention Module(CCBAM). The motive and design details of the above modules are presented below.

In detail, to overcome the large number of parameters and slow inference speed caused by the multi-path encoder, we employ Short-Term Dense Concatenate network-STDC\cite{fan2021rethinking} derived from DenseNet\cite{hung2019incorporating} as our backbone. In order to obtain variable receptive fields with fewer parameters, the number of filters of convolutional layers in the STDC blocks is gradually reduced. The STDC module concatenates response maps from multiple consecutive layers, each encoding input images/features at different scales and respective domains, resulting in a multi-scale feature representation.
Enhanced feature representation plays a crucial role in improving semantic segmentation accuracy\cite{li2020semantic,song2021attanet}. It is usually achieved by fusing the output features from the low-level and high-level of the decoder. However, the existing fusion models are too complex, resulting in a huge amount of model parameters. In this work, we propose a Cross Convolutional Block Attention Module(CCBAM) to strengthen feature representations efficiently. As shown in Figure \ref{fig3_Cross-cbam}, CCBAM first utilizes a cross-attention module for fusing features. It uses both channel attention and spatial attention, channel attention emphasizes 'what' while spatial attention emphasizes 'where'.

Variable receptive fields and multi-scale information are key factors to improve semantic segmentation accuracy. Existing methods such as parallel Atrous Spatial Pyramid Pooling module (ASPP)\cite{chen2017rethinking} adopt four parallel atrous convolutions to obtain multi-scales. It expands the receptive fields while increasing parameters, which is time-consuming for real-time networks. Therefore, we design an innovative and lightweight module called the Squeeze-and-Excitation Atrous Spatial Pyramid Pooling Module(SE-ASPP) to expand the field of view of filter. SE-ASPP only keeps two parallel atrous convolutions, and replaces the concatenate operation with an add operation. We replace image pooling in ASPP with a SE attention module.

Our main contributions can be summarized as follows:
\begin{itemize}
    \item A new feature fusion module called CCBAM is designed. It boasts the cross-multiplication operation that enables high-level semantic information to guide low-level details information. And it leverages channel and spatial attention to strengthen the feature representations.
    \item A Squeeze-and-Excitation Atrous Spatial Pyramid Pooling Module(SE-ASPP) is proposed, which can not only obtain a larger and variable receptive field, but also multiple scales. While SE-ASPP reduces the complexity of the model and ensures the accuracy of the segmentation.
    \item A lightweight attention segmentation network is designed for the real-time semantic segmentation task network. Cross-CBAM utilizes contextual attention fusion module(CCBAM) to fuse information, and utilizes SE-ASPP to further improve the accuracy at a very small cost of speed. 
    \item A series of experiments demonstrate the superiority of the proposed method. Experimental results show that the Cross-CBAM network achieves a good trade-off between segmentation accuracy and inference speed. Specifically, Cross-CBAM-L1 network achieves $75.1\%$ mIoU on the Cityscapes test set at a speed of $187.9$FPS. Cross-CBAM-L2 achieves $77.2\%$ mIoU and $88.6$FPS under the same experimental setting.
\end{itemize}

\section{Related work}
\label{sec:formatting}

\subsection{Semantic Segmentation}
Traditional semantic segmentation algorithms assign labels to each pixel using handcrafted features. With the development of deep convolution neural networks, semantic segmentation has achieved outstanding performance. FCN\cite{long2015fully} is the first full convolutional network for semantic segmentation. Due to its fully convolutional nature, it can segment any input image resolution. Furthermore, it is trained in an end-to-end and pixel-to-pixel manner. SegNet\cite{badrinarayanan2017segnet} proposed that the decoder perform non-linear upsampling using max-pooled indices. DeepLabv3\cite{chen2017rethinking} adopted a cascade or parallel ASPP module to capture multi-scale context by using multiple atrous rates. PSPNet\cite{zhao2017pyramid} exploits the global contextual information with contextual aggregation based on different regions. However, most of these methods require substantial computational cost due to the high resolution of images. In this paper, we propose an efficient and effective segmentation model that achieves a good trade-off between speed and accuracy.

\begin{figure}
    \includegraphics[width=0.5\textwidth]{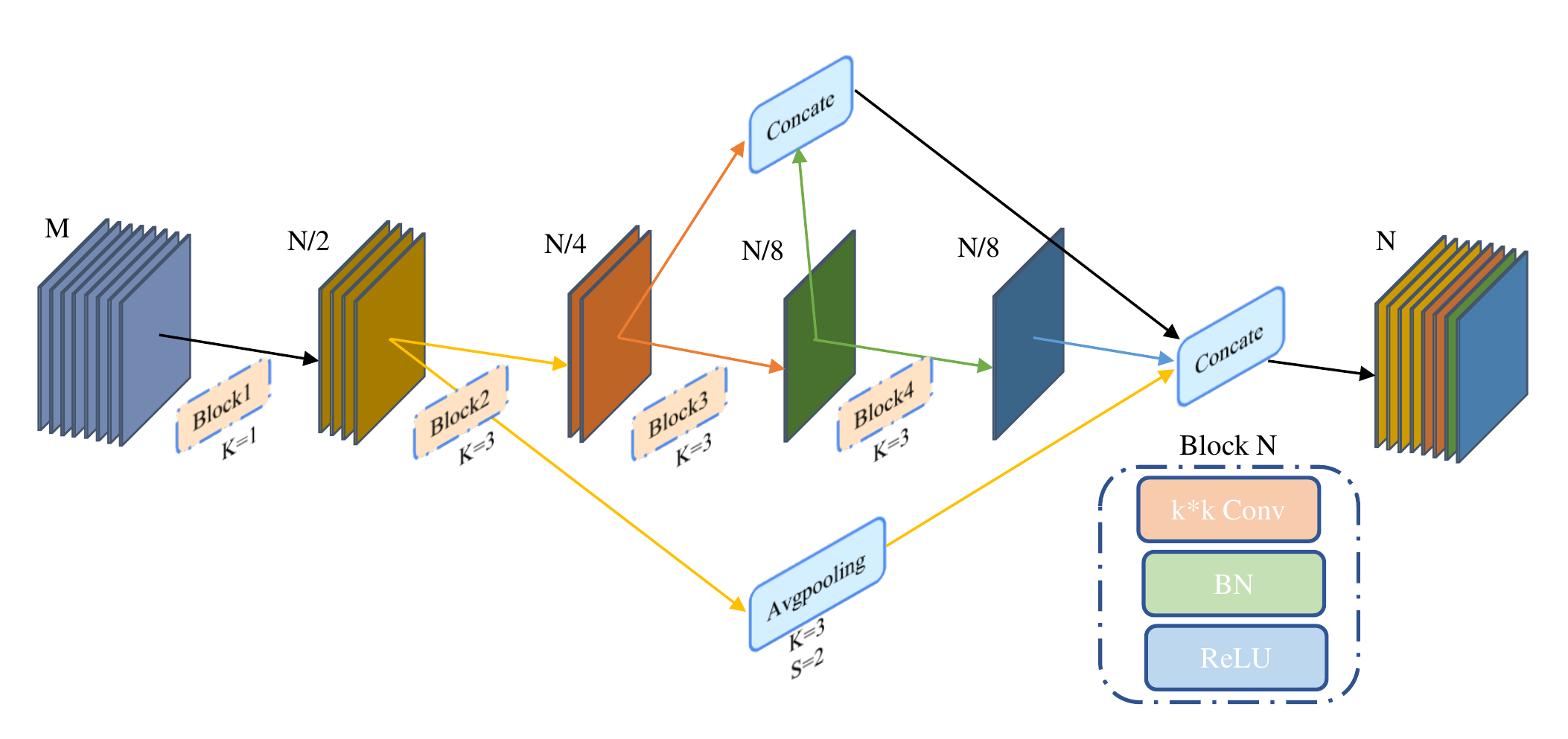}
    \caption{\textbf{The STDC module used in our network.} \textit{Block i} operation includes Conv-BN-ReLU, \textit{Avgpooling} operation refers to average pooling with stride=2 and kernel-size=3. \textit{M} denotes the dimension of the input channels, and \textit{N} denotes the dimension of the output channels.}
    \label{fig2_STDC module}
\end{figure}

\subsection{Real-Time Semantic Segmentation}
Due to the limitation of computing resources, traditional semantic segmentation methods are greatly challenged. Therefore, lightweight semantic segmentation models are urgently needed to solve practical application problems. There are two mainstream approaches to designing efficient models. (1)\textit{lightweight backbone}. MobileNet series\cite{howard2017mobilenets,sandler2018mobilenetv2,howard2019searching} replaces traditional convolutions with depthwise separable convolutions. The depthwise separable convolution is composed of two parts: depthwise (DW) and pointwise (PW), and its number of parameters and operation cost are lower than traditional convolution. ShuffleNet\cite{zhang2018shufflenet} utilizes Pointwise Group Convolution and channel shuffle to reduce network capacity. (2)\textit{lightweight decoder}. LEDNet\cite{wang2019lednet} designs an attention pyramid network (APN) in the decoder to alleviate the complexity of the network. PP-LiteSeg\cite{peng2022pp} proposes a flexible and lightweight decoder (FLD), which reduces the computational load of the decoder.

\begin{figure}
    \includegraphics[width=0.5\textwidth]{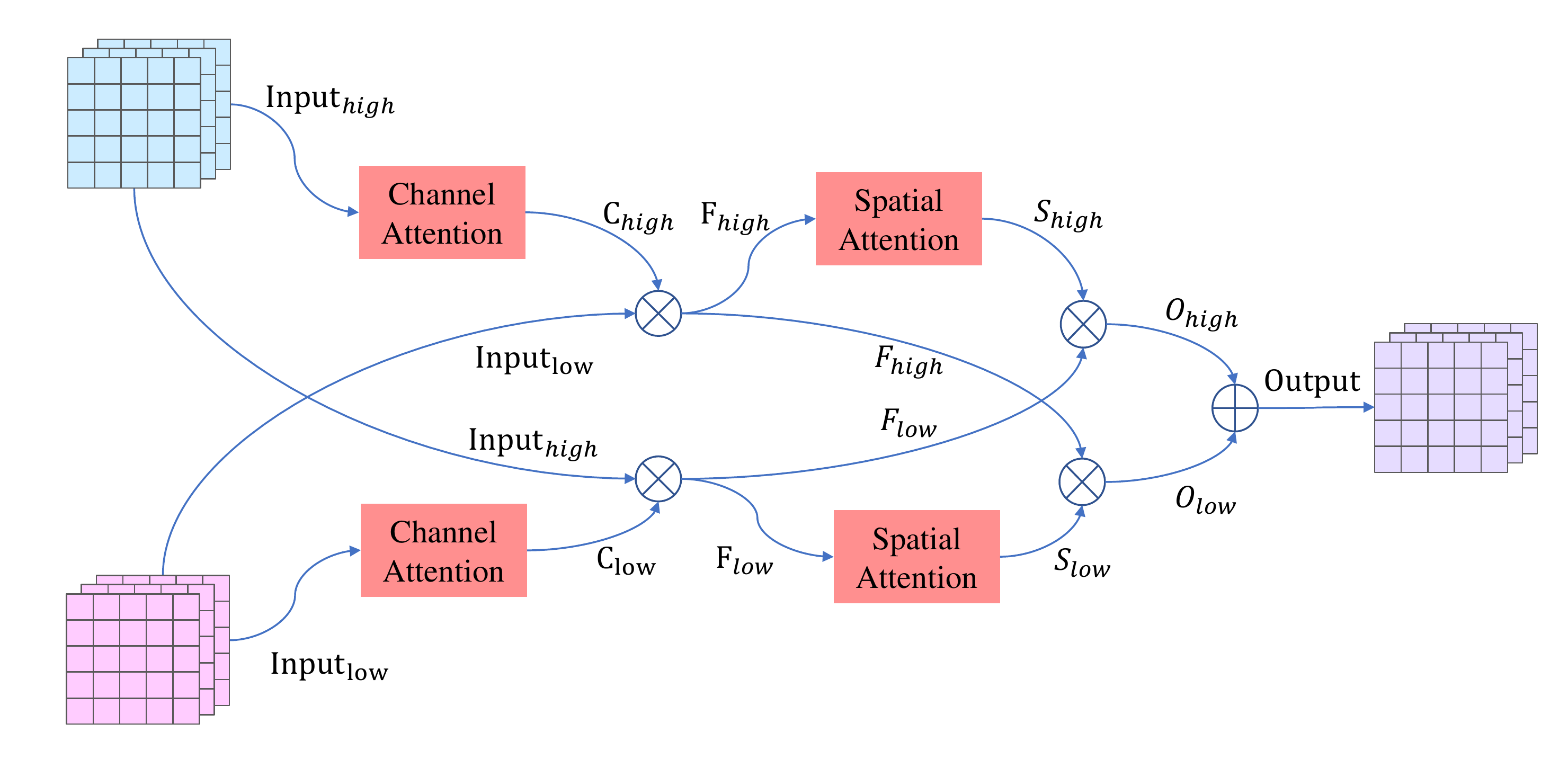}
    \caption{\textbf{Diagram of CCBAM.} We multiply the feature response from low-level with the feature from high-level, and multiply the feature response from high-level with the feature from low-level. Then, we add the obtained features to get the final output.}
    \label{fig3_Cross-cbam}
\end{figure}
\begin{figure}
    \includegraphics[width=0.5\textwidth]{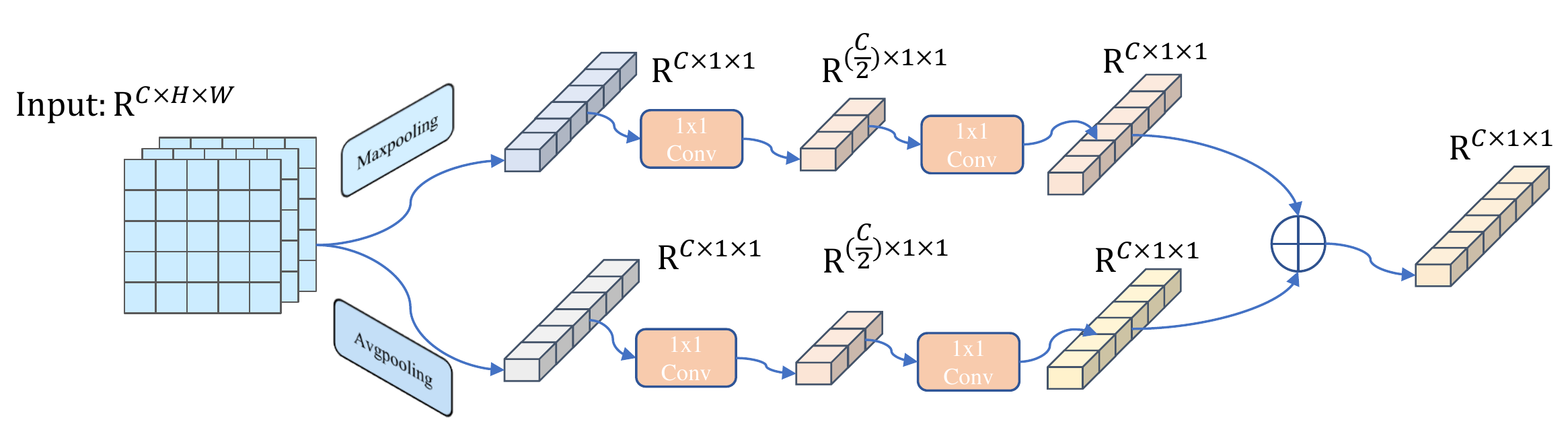}
    \caption{\textbf{Diagram of Channel Attention Module.}}
    \label{fig4_Channels attention module}
\end{figure}
\begin{figure}
    \includegraphics[width=0.5\textwidth]{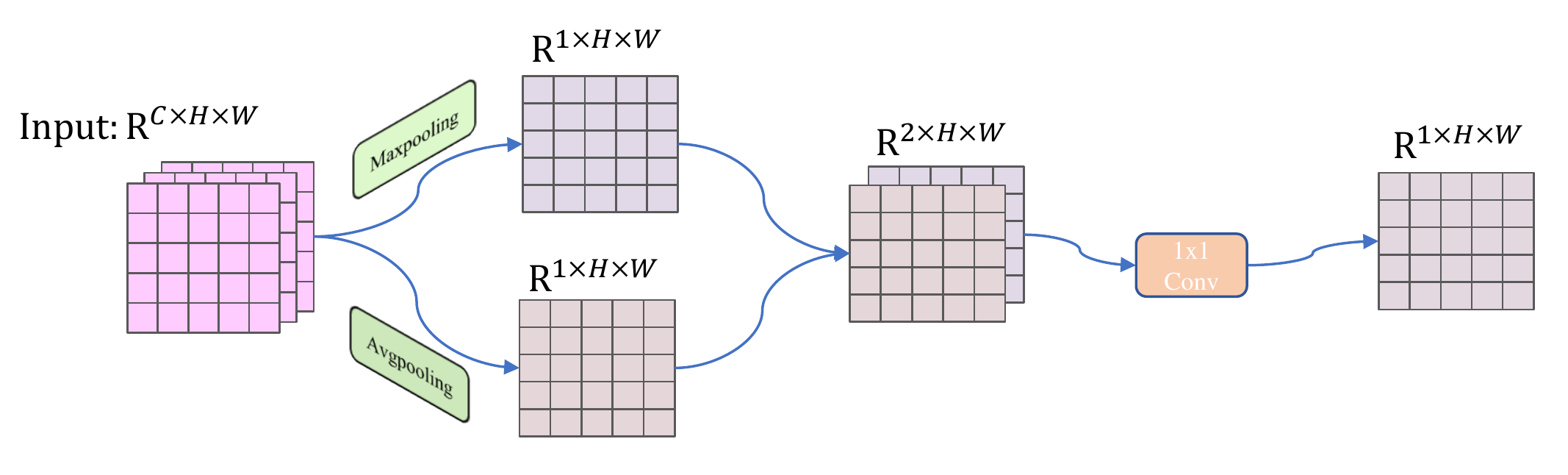}
    \caption{\textbf{Diagram of Spatial Attention Module.}}
    \label{fig5_Spatial attention module}
\end{figure}

\subsection{Attention mechanism}
Attention can be interpreted as a means of selectively focusing on a portion of all information. Squeeze-and-Excitation Network\cite{hu2018squeeze}(SENet) uses global average pooling features to calculate channel-wise attention. It reweights feature channels to recalibrate channel-wise feature responses. Gather-Excite: Exploiting Feature Context in Convolutional Neural Networks \cite{hu2018gather}(GENet) introduces a pair of operators: gather and excite. 'Gather'  aggregates feature responses that are from a large spatial extent efficiently, and 'excite' redistributes the pooled information to local features. Convolutional Block Attention Module\cite{woo2018cbam}(CBAM) uses both channel attention and spatial attention to infer attention maps. The attention maps are then multiplied by the input feature map for feature refinement.

\section{Proposed Method}
\label{sec:propose}
In this section, we first introduce the Short-Term Dense Concatenate Module(STDC). Then Cross Convolutional Block Attention Module(CCBAM) and Squeeze-and-Excitation Atrous Spatial Pyramid Pooling Module(SE-ASPP) are recommended respectively. Finally, we show the whole structure of our real-time semantic segmentation model.

\begin{figure}
    \includegraphics[width=0.5\textwidth]{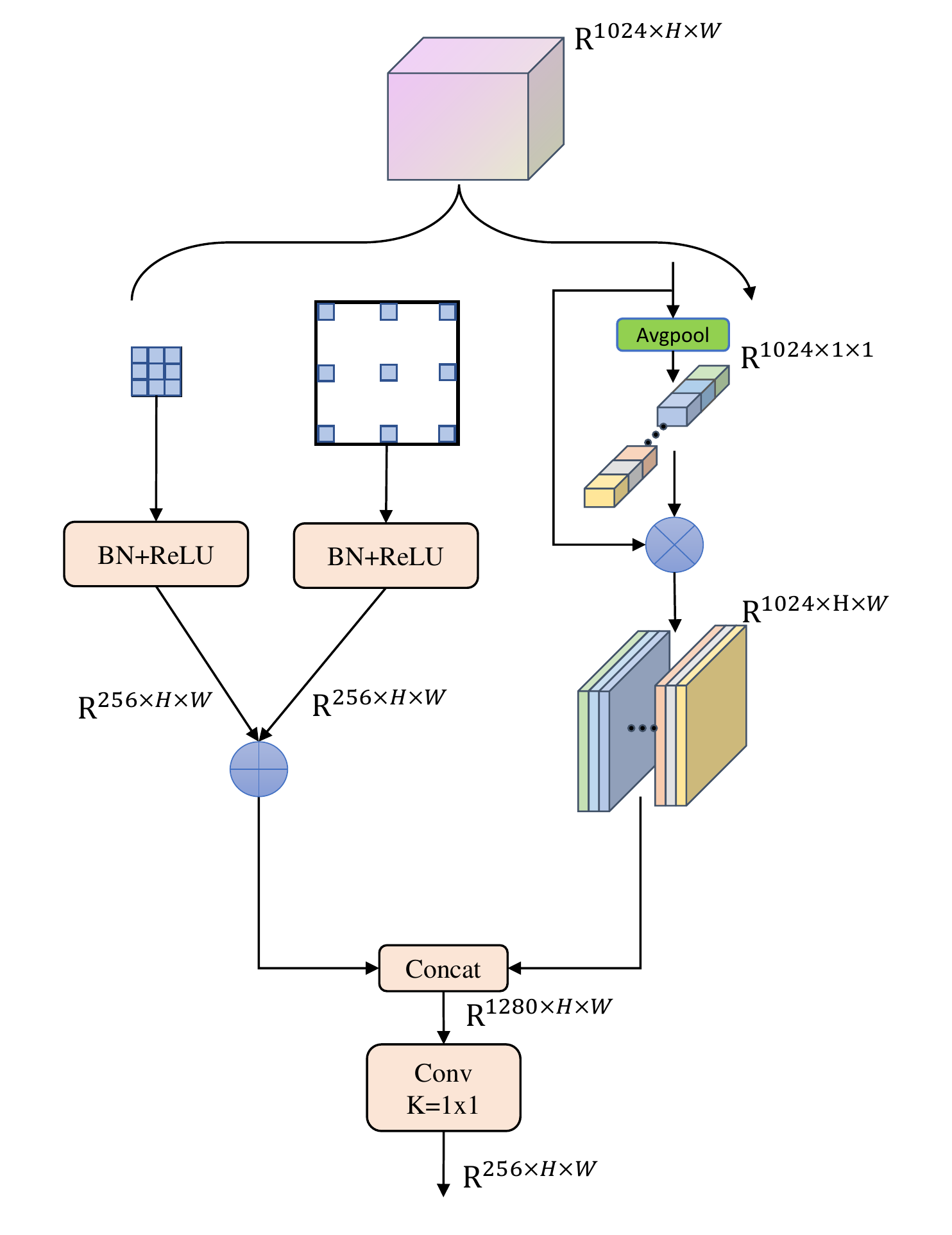}
    \caption{\textbf{Details of the proposed SE-ASPP module.} We adopt atrous convolution with dilation-rates=$(1,3)$, at the same time, the dilation rates can be adjusted dynamically.}
    \label{fig6_SE-ASPP module}
\end{figure}

\subsection{Short-Term Dense Concatenate Module}
Typically, an encoder is designed to extract hierarchical features, the encoder consists of a series of layers grouped into several stages. As the number of convolutional neural network layers increases from shallow to deep, the number of channels gradually increases, and the spatial resolution gradually decreases. The Short-Term Dense Concatenate network\cite{fan2021rethinking} also has the same structure. Due to a large number of parameters, STDC abandoned the dual-stream path proposed by Bisenet\cite{yu2018bisenet} and switched to a single-path decoder. Due to the Short-Term Dense Concatenate structure and the design of Detail Guidance, it achieves $71.9\%$ mIoU and $250.4$FPS on the Cityscapes test set, which is an excellent trade-off between accuracy and speed. Figure \ref{fig2_STDC module} illustrates the layout of the STDC module. Specifically, the STDC module is divided into several blocks, and each block includes one convolutional layer, one BN layer and a ReLU activation layer. The first block in the STDC module has a kernel size of 1, and the remaining blocks have a kernel size of 3. Assuming that the number of output channels of the STDC module are N, the number of output channels of the $i$-th block is $\frac{N}{2^i}$, except for the number of channels of the last block. It has the same number of output channels as the previous block. To enrich feature information, the STDC module concatenates the output feature maps of each block as the output of the STDC module through a skip path. Before concatenation, the feature maps from different blocks in the STDC module are down-sampled to the same spatial resolution by average-pooling operation with a pool size of 3 $\times$ 3. The final output of the STDC module is:
\begin{equation}
   x_{\text {output }}=F\left(x_{1}, x_{2}, \ldots, x_{n}\right) 
\end{equation}
where $x_{\text {output }}$ denotes the output of the STDC module, F denotes concatenate operation, and $x_{1}, x_{2}, \ldots, x_{n}$ denote the feature maps from all blocks. 

\subsection{Cross Convolutional Block Attention Module}
Feature fusion plays a critical role in feature representation. Commonly used feature representation methods are element-wise summation and concatenation. But in this paper, we propose a new attention-based feature fusion module called CCBAM, which utilizes both channel attention and spatial attention. Different from the previous position of attention, CCBAM uses cross attention in the feature fusion of Encoder and decoder.
\paragraph{CCBAM Framework.} As shown in figure \ref{fig3_Cross-cbam}, CCBAM has two input feature maps, and after passing through the channel attention module and the spatial attention module, the two inputs are fused into one output. In detail, the two input feature maps are denoted as $ Input_{high} $ and $ Input_{low} $. $ Input_{high} $ is the output feature map of the decoder, and $ Input_{low} $ is the counterpart of the decoder. $ Input_{high} $ and $ Input_{low} $ go through the channel attention module and generate channel-wise feature responses $C_{high}$ and $C_{low}$, respectively. Then, We multiply $ Input_{low} $ and $C_{high}$ to get $F_{high}$, multiply $ Input_{high} $ and $C_{low}$ to get $F_{low}$. $F_{high}$ and $F_{low}$ go through the spatial attention module and generate spatial-wise feature responses $S_{high}$ and $S_{low}$, respectively. After that, we multiply $F_{low}$ and $S_{high}$ to get the $O_{high}$, and multiply $F_{high}$ and $S_{low}$ to get the $O_{low}$. Finally, we apply element-wise operation to add $O_{high}$ and $O_{low}$ to get $Output$. Note that $O_{high}$ and $O_{low}$ have the same channels. As shown above, the designed module is called CCBAM since we cross-multiply feature maps and feature responses. We formulate the above procedure as equation \ref{equation_2}.
\begin{equation}
\begin{split}
     C_{high} = Channel\quad Attention (Input_{high})
     \\
     C_{low} = Channel \quad Attention (Input_{low})
     \\
     F_{high} = Input_{low} \cdot C_{high}
     \\
     F_{low} = Input_{high} \cdot C_{low}
     \\
     S_{high} = Spatial \quad Attention(F_{high})
     \\
     S_{low} = Spatial \quad Attention(F_{low})
     \\
     Output = ((F_{low} \cdot S_{high})+(F_{high} \cdot S_{low}))
     \label{equation_2}
\end{split}
\end{equation}
\paragraph{Channel Attention Module.}
In the channel attention module, both average pooling and max pooling operations are used to aggregate spatial information of a feature map. As shown in Figure \ref{fig4_Channels attention module}, the size of the feature map for a given input is $R^{C\times H \times W}$. After max-pooling and average-pooling operations, two feature outputs will be generated, namely average-pooled feature and max-pooled features, and the dimension of each feature is $R^{C\times 1 \times 1}$. Then, convolution is performed with one hidden layer operation, and the hidden layer activation size is set to $R^{\frac{C}{r}\times 1 \times 1}$, where r is the reduction ratio(r=16 in our method). Finally, we combine the output feature maps with an element-wise summation. The procedure can be formulated as equation \ref{equation_3}.

\begin{equation}
\begin{aligned}
     C=Conv(Maxpooling(Input))+
    \\Conv(Avgpooing(Input))
    \label{equation_3}
\end{aligned}
\end{equation}

\paragraph{Spatial Attention Module.}
Different from channel attention, which focuses on 'what', spatial attention focuses on 'where'. As shown in Figure \ref{fig5_Spatial attention module}, given the input features $ I^{C\times H\times W} $, first apply max pooling and average pooling along the channel axis to generate two 2D feature response maps, of which the dimension is $ R^{1\times H\times W} $. A concatenation operation is then applied to the two feature response maps and produces an effective feature descriptor, the dimension is $ R^{2\times H\times W} $. Afterwards, convolution and sigmoid operations are applied to generate a spatial attention map of dimension $ R^{1\times H\times W} $. Note that in CBAM, the kernel size of the last convolution operation is $7\times 7$, but only a $1\times 1$ kernel is used for channel transformation in this work. The procedure can be formulated as equation \ref{equation_4}.
\begin{equation}
\begin{aligned}
    S=Conv(Concatenate(Maxpooling(Input),
    \\Avgpooling(Input)))
    \label{equation_4}
\end{aligned}
\end{equation}

\begin{table}[]
\caption{Details of the STDC networks. \textit{ConvX} stands Conv-BN-ReLU. \textit{KSize} means kernel size. \textit{S, R, C} denote stride, the number of repetitions, and output channels respectively.}\label{STDC netwoks}
\setlength{\tabcolsep}{1.8mm}{
\begin{tabular}{c|c|c|c|cc|cc}
\Xhline{1pt}
\multirow{2}{*}{Stages} & \multirow{2}{*}{Output size}                          & \multirow{2}{*}{KSize} & \multirow{2}{*}{S}                            & \multicolumn{2}{c|}{STDC1}                                                & \multicolumn{2}{c}{STDC2}                                                 \\ \cline{5-8} 
                        &                                                       &                        &                                               & \multicolumn{1}{c|}{R}                                             & C    & \multicolumn{1}{c|}{R}                                             & C    \\ \hline
Image                   & 224$\times$224                                               &                        &                                               & \multicolumn{1}{c|}{}                                              & 3    & \multicolumn{1}{c|}{}                                              & 3    \\ \hline
ConvX1                  & 112$\times$112                                               & 3x3                    & 2                                             & \multicolumn{1}{c|}{1}                                             & 32   & \multicolumn{1}{c|}{1}                                             & 32   \\ \hline
ConvX2                  & 56$\times$56                                                 & 3x3                    & 2                                             & \multicolumn{1}{c|}{1}                                             & 64   & \multicolumn{1}{c|}{1}                                             & 64   \\ \hline
Stage3                  & \begin{tabular}[c]{@{}c@{}}28$\times$28\\ 28$\times$28\end{tabular} &                        & \begin{tabular}[c]{@{}c@{}}2\\ 1\end{tabular} & \multicolumn{1}{c|}{\begin{tabular}[c]{@{}c@{}}1\\ 1\end{tabular}} & 256  & \multicolumn{1}{c|}{\begin{tabular}[c]{@{}c@{}}1\\ 3\end{tabular}} & 256  \\ \hline
Stage4                  & \begin{tabular}[c]{@{}c@{}}14$\times$14\\ 14$\times$14\end{tabular} &                        & \begin{tabular}[c]{@{}c@{}}2\\ 1\end{tabular} & \multicolumn{1}{c|}{\begin{tabular}[c]{@{}c@{}}1\\ 1\end{tabular}} & 512  & \multicolumn{1}{c|}{\begin{tabular}[c]{@{}c@{}}1\\ 4\end{tabular}} & 512  \\ \hline
Stage5                  & \begin{tabular}[c]{@{}c@{}}7$\times$7\\ 7$\times$7\end{tabular}     &                        & \begin{tabular}[c]{@{}c@{}}2\\ 1\end{tabular} & \multicolumn{1}{c|}{\begin{tabular}[c]{@{}c@{}}1\\ 1\end{tabular}} & 1024 & \multicolumn{1}{c|}{\begin{tabular}[c]{@{}c@{}}1\\ 2\end{tabular}} & 1024 \\
\Xhline{1pt}
\end{tabular}}
\end{table}

\subsection{Squeeze-and-Excitation Atrous Spatial Pyramid Pooling Module}
Atrous Spatial Pyramid Pooling Module has been proven to be effective in semantic segmentation tasks\cite{chen2017rethinking}. The atrous rates adopted by the ASPP module are $ 1,12,24,36$, which is of great help to effectively capture multi-scale information. However, large atrous rates, four parallel atrous convolutions and bilinear interpolation bring a huge amount of parameters and computational costs, which are not suitable for real-time semantic segmentation networks. Therefore, we design a Squeeze-and-Excitation Atrous Spatial Pyramid Pooling Module(SE-ASPP module), which has fewer parameters and faster inference speed for real-time. As shown in Figure \ref{fig6_SE-ASPP module}. SE-ASPP consists of two parallel atrous convolutions and a SE attention module, one of which is $1\times 1 $ convolution with $rate=1$, and the other is $3\times 3 $ convolution with $rate=3$(All atrous convolution with 256 filters and batch normalization, and experiments show that 256 channels achieved trade-off between accuracy and speed). In addition, the bilinear interpolation upsampling in the ASPP module is computationally expensive, so the bilinear interpolation is discarded and the SE attention module is utilized. The resulting features from atrous convolution are added and then concatenate the resulting features from SE module. Finally, a convolutional layer is adopted for channel transformation.

\begin{table}[]
\caption{\label{comparison with state of the art} The comparisons with other state-of-the-art methods on Cityscapes. \textit{no} means the method has no backbone. " - " indicates that the method does not give official data in their paper.}
\setlength{\tabcolsep}{0.5mm}{
\begin{tabular*}{\linewidth}{llllll}
\toprule
\multirow{2}{*}{Model} & \multicolumn{1}{l}{\multirow{2}{*}{Backbone}} & \multicolumn{1}{l}{\multirow{2}{*}{Resolution}} & \multicolumn{2}{l}{mIoU(\%)}                          & \multirow{2}{*}{FPS} \\ 
                                & \multicolumn{1}{l}{}                                   & \multicolumn{1}{l}{}                                     & \multicolumn{1}{l}{val}            & test                      &                               \\ 
\midrule
ENet\cite{paszke2016enet}                                                 & no                                 & 512x1024                             & \multicolumn{1}{l}{-}              & 58.3          & 76.9                          \\
LDFNet\cite{hung2019incorporating}                                              & no                                 & 512x1024                             & \multicolumn{1}{l}{68.4}           & 71.3          & 27.7                          \\
BIseNetV2-L\cite{yu2021bisenet}                                          & no                                 & 512x1024                             & \multicolumn{1}{l}{75.8}           & 75.3          & 47.3                          \\
CAS\cite{zhang2019customizable}                                                  & no                                 & 768x1536                             & \multicolumn{1}{l}{-}              & 70.5          & 108.0                         \\
GAS\cite{lin2020graph}                                                  & no                                 & 769x1537                             & \multicolumn{1}{l}{-}              & 71.8          & 108.4                         \\
HMSeg\cite{li2020humans}                                                & no                                 & 768x1536                             & \multicolumn{1}{l}{-}              & 74.3          & 83.2                          \\
FasterSeg\cite{chen2019fasterseg}                                            & no                                 & 1024x2048                            & \multicolumn{1}{l}{73.1}           & 71.5          & 163.9                         \\
MFNet\cite{ha2017mfnet}                                    & no                      & 512x1024                                                        & \multicolumn{1}{l}{-}           & 72.1          & 116                             \\
LSPANet\cite{xiao2022real}                                    & no                       & 1024x2048                            & \multicolumn{1}{l}{-}           & 77.1          & 60.3                             \\
SegFormer\cite{xie2021segformer}                                    & no                      & 1024x1024                            & \multicolumn{1}{l}{-}           & 76.2          & 15.2                             \\
SETR-PUP\cite{zheng2021rethinking}                                    & no                      & 768x768                            & \multicolumn{1}{l}{-}           & 78.4          & -                             \\
EDANet\cite{lo2019efficient}                                              & EDANet                             & 512x1024                             & \multicolumn{1}{l}{-}              & 67.3          & 108.7                         \\
ESPNetV2\cite{mehta2019espnetv2}                                             & ESPV2                              & 512x1024                             & \multicolumn{1}{l}{66.4}           & 66.2          & -                             \\
LiteSeg\cite{emara2019liteseg}                                              & ShuffleNet                         & 512x1024                             & \multicolumn{1}{l}{67.8}           & -             & 133.0                           \\
SFNet\cite{li2020semantic}                                                & DF1                                & 1024x2048                            & \multicolumn{1}{l}{-}              & 74.5          & 121.0                          \\
ICNet\cite{zhao2018icnet}                                                & PSPNet50                           & 1024x2048                            & \multicolumn{1}{l}{-}              & 69.5          & 30.3                          \\
MobilenetV3-L\cite{howard2019searching}                                    & MobiletnetV3                       & 1024x2048                            & \multicolumn{1}{l}{72.6}           & 72.6          & -                             \\
SegNext-T\cite{howard2019searching}                                    & MSCAN-T                      & 768x1536                            & \multicolumn{1}{l}{-}           & 78.0          & 25.0                             \\
STDC1-Seg50\cite{fan2021rethinking}                                          & STDC1                              & 512x1024                             & \multicolumn{1}{l}{72.2}           & 71.9          & 250.4                         \\
STDC2-Seg50\cite{fan2021rethinking}                                          & STDC2                              & 512x1024                             & \multicolumn{1}{l}{74.2}           & 73.4          & 188.6                         \\
STDC1-Seg75\cite{fan2021rethinking}                                          & STDC1                              & 768x1536                             & \multicolumn{1}{l}{74.5}           & 75.3          & 126.7                         \\
STDC2-Seg75\cite{fan2021rethinking}                                          & STDC2                              & 768x1536                             & \multicolumn{1}{l}{77.0}             & 76.8          & 97.0                          \\ 
PP-LiteSeg-T1\cite{peng2022pp}                                          & STDC1                              & 512x1024                             & \multicolumn{1}{l}{73.1}           & 72          & 273.6                         \\
PP-LiteSeg-B1\cite{peng2022pp}                                          & STDC2                              & 512x1024                             & \multicolumn{1}{l}{75.3}           & 73.9          & 195.3                         \\
PP-LiteSeg-T2\cite{peng2022pp}                                            & STDC1                              & 768x1536                             & \multicolumn{1}{l}{76.0}           & 74.9          & 143.6                         \\
PP-LiteSeg-B2\cite{peng2022pp}                                            & STDC2                              & 768x1536                             & \multicolumn{1}{l}{78.2}             & 77.5          & 102.6                         \\ 
\midrule
Cross-CBAM-M1                                        & STDC1                              & 512x1024                             & \multicolumn{1}{l}{74.19}          & 73.4          & 240.9                         \\
Cross-CBAM-L1                                        & STDC2                              & 512x1024                             & \multicolumn{1}{l}{76.04}          & 75.1          & 187.9                         \\
Cross-CBAM-M2                                        & STDC1                              & 768x1536                             & \multicolumn{1}{l}{76.14}          & 75.5          & 119.7                         \\
Cross-CBAM-L2                                        & STDC2                              & 768x1536                             & \multicolumn{1}{l}{78.35} & 77.2 & 88.6\\                            
\bottomrule
\end{tabular*}}
\end{table}

\begin{table}[htbp]
\caption{The comparisons with other state-of-the-art methods on CamVid. \textit{no} means the method has no backbone. " - " indicates that the method does not give official data in their paper.}\label{camvid-comparison with state of the art} 
\begin{tabular*}{\tblwidth}{@{}LLLL@{}}
\toprule
 Model                    & Backbone  & mIoU(\%) & FPS   \\ 
\midrule
ENet\cite{paszke2016enet}                   & no        & 51.3     & 61.2  \\ 
CAS\cite{zhang2019customizable}             & no        & 71.2     & 169  \\ 
GAS\cite{lin2020graph}                      & no        & 72.8     & 153.1  \\ 
EDANet\cite{lo2019efficient}                & no        & 66.4     & -     \\ 
BiseNetV2-L\cite{yu2021bisenet}             & no        & 73.2     & 32.7  \\ 
LSPANet\cite{xiao2022real}                  & no        & 73.0     & 73.5  \\ 
MFNet\cite{ha2017mfnet}                     & no        & 71.5     & 145  \\ 
DDRNet\cite{hong2021deep}                   & ResNet    & 74.7     & 230  \\
BiseNetV1-L\cite{yu2018bisenet}             & ResNet18  & 68.7     & 116.3 \\ 
ICNet\cite{zhao2018icnet}                   & PSPNet50  & 67.1     & 34.5  \\ 
TD4-PSP18\cite{hu2020temporally}            & PSPNet-18 & 72.6     & 40    \\ 
STDC1-Seg\cite{fan2021rethinking}           & STDC1     & 73.0     & 197.6 \\ 
STDC2-Seg\cite{fan2021rethinking}           & STDC2     & 73.9     & 152.2 \\ 
PP-LiteSeg-T\cite{peng2022pp}               & STDC1     & 73.3     & 222.3 \\ 
PP-LiteSeg-B\cite{peng2022pp}               & STDC2     & 75.0     & 154.8 \\ 
\midrule
Cross-CBAM-M                                & STDC1     &  73.8              & 185.3      \\ 
Cross-CBAM-L                                & STDC2     & \textbf{75.6}      & 146.8      \\
\bottomrule
\end{tabular*}
\end{table}
\begin{table}[<options>]
\caption{Ablation experiments of our proposed model on the Cityscapes validation set. }\label{ablation study} 
\begin{tabular*}{\tblwidth}{@{}LLLLL@{}}
\toprule
 Model         & SE-ASPP & Cross-CBAM & Aux-loss & mIoU(\%) \\
\midrule
Baseline      &         &            &           & 52.26    \\
Cross-CBAM-M1 &    $\checkmark$     &            &           & 68.74    \\
Cross-CBAM-M1 &         &     $\checkmark$       &           & 72.63    \\
Cross-CBAM-M1 &    $\checkmark$     &      $\checkmark$      &           & 74.02    \\
Cross-CBAM-M1 &     $\checkmark$    &    $\checkmark$        &     $\checkmark$      & 74.19    \\ 
\bottomrule
\end{tabular*}
\end{table}

\subsection{Network Architecture}
The proposed network structure is demonstrated in Figure \ref{fig1_overview_of_crosscbam}. In the decoder, the STDC network\cite{fan2021rethinking} is chosen to extract hierarchical information. It has 5 stages, each of which downsamples the input with a stride of 2 to twice resolution, so that the final output feature map is only 1/32 of the size of the original image. Table \ref{STDC netwoks} shows the details of the STDC network, only the output of stages 3,4,5 are used for feature fusion. This work proposes two versions of the network: Cross-CBAM-M and Cross-CBAM-L, where the decoders are STDC1\cite{fan2021rethinking} and STDC2\cite{fan2021rethinking} respectively. Cross-CBAM-M achieves faster inference speed, while Cross-CBAM-L achieves higher segmentation accuracy. First, we pass the output of the last layer of the encoder, that is, the output of stage 5, through the SE-ASPP module. A feature map is received, which has 256 channels and is 1/32 the size of the original image. Afterwards, an upsampling operation is applied to resize these feature maps to the size of the stage 4 output feature maps. Then, the CCBAM module is utilized to fuse the output from stage 4 and the resized feature maps. Next, we resize the output feature maps of Cross-CBAM to the size of the stage 3 output feature map. Finally, the CCBAM module is used to fuse the output feature maps of stage 3 and the resized feature maps. 

In addition, due to the long tail problem of various data, an additional focal loss is adopted \cite{lin2017focal} to alleviate this problem. Focal loss was originally designed to address the one-stage object detection scenario and class imbalance. It originates from the cross entropy loss for binary classification, equation \ref{eqution_5}. 
\begin{equation}
\begin{aligned}
\mathrm{CE}(p, y)= \begin{cases}-\log (p) & \text { if } y=1 \\ -\log (1-p) & \text { otherwise }\end{cases}
\label{eqution_5}
\end{aligned}
\end{equation}
where $y \in \{\pm 1\}$ specifies the ground-truth class and $p \in[0,1] $ is the model’s estimated probability for the class with label $y = 1$. For convenience, it rewrites $CE(p,y)=CE(P_t)=-log(p_t)$, $p_t=p$ if $y=1$, else $p_t=1-p$. Focal loss adds a modulation factor $(1-p_t)^{\gamma}$ to cross entropy loss, so as to remodel the loss function, reduce the weight of large scale targets, and focus on small targets training. It can be defined as:
\begin{equation}
    FL(p_t)=-(1-p_t)^\gamma \log(p_t)
\end{equation}
where $\gamma \in [0,5]$ is a tunable parameter.

At the same time, with the increasing depth of the network, the parameters in the network are difficult to optimize, and the auxiliary loss can help optimize the learning process. In order to balance the main loss and auxiliary loss, a weight $\alpha$ is set to choose the proportion of main loss and auxiliary loss. The loss function in this paper is designed as: $Loss=\alpha CE + (1-\alpha)FL$.  Note that the auxiliary loss is only used during training, and during the testing phase, the auxiliary loss is abandoned and only the main branch is used.

\begin{table}[<options>]
\caption{Atrous Rates Ablation Study. As can be seen from the table, dilation=(1,3) achieves a trade-off between speed and accuracy. \textit{Note that we set the number of channels of SE-ASPP to 256 and the input size to $512\times1024$.}}\label{atrous rates} 
\begin{tabular*}{\tblwidth}{@{}LLLLL@{}}
\toprule
 Dilations & mIoU           & FPS               & Flops(G)                      & Parameters(M)         \\
\midrule
(1,3)     & 74.19           & \textbf{241.1}   & \textbf{11.31}                 & \textbf{12.21} \\
(2,4)     & 74.20           &   229.5          &   12.39                        &    14.31                      \\
(3,5)     &  75.30          &  229.4           & 12.39                          &       14.31                     \\
(1,3,5)   & 74.61           & 226.4           & 12.52                          &        14.57                   \\
(2,4,6)      & 74.15              & 225.1              & 13.60                                  &16.67                   \\
\bottomrule
\end{tabular*}
\end{table}

\begin{table}[<options>]
\caption{Comparisons with different channels on Cityscapes. \textit{Note that we set the atrous rates to (1,3). The Input size is set to $512\times1024$.}}\label{se-aspp channels} 
\begin{tabular*}{\tblwidth}{@{}LLLLL@{}}
\toprule
Channels & mIoU           & FPS   & Flops(G)           & Parameters(M) \\
\midrule
512      & 73.86          &235.7    & 14.77            & 19.54         \\
256      & \textbf{74.19} & 240.9   & 11.31            & 12.21         \\
128      & 72.21          &245.6    &\textbf{9.67}     &\textbf{8.92}           \\
\bottomrule
\end{tabular*}
\end{table}

\section{Experiments Results}
In this section, we first introduce the datasets used in this paper and the implementation details. Then, we compare the experimental results with existing models in terms of accuracy and inference speed. Finally, extensive ablation experiments are conducted to demonstrate the effectiveness of the proposed module.
\subsection{Benchmarks and Evaluation Metrics}
\paragraph{Cityscapes.}The Cityscapes\cite{Cordts2016} is taken from the perspective of cars and is one of the well-known datasets focusing on the parsing of urban streetscape. It contains multiple stereoscopic video sequences recorded in street scenes from 50 different cities, with 5000 high-quality pixel-level annotations in addition to 20000 coarsely annotated frames. We only used 5,000 fine annotated images, 2,975 for training, 500 for validation, and 1525 for testing. Annotations include 30 classes, 19 of which are used for semantic segmentation tasks. Furthermore, the resolution of the image is as high as $1024\times2048$, which is a great challenge for real-time semantic segmentation tasks.

\paragraph{CamVid.}The CamVid\cite{brostow2009semantic} is taken from the perspective of a driving automobile, which is for road scene parsing. The database provides ground truth labels associating each pixel with one of 32 semantic classes, a subset of 11 classes are used for semantic segmentation experiment. The dataset contains 701 images with high-quality annotations, of which 367 images are used for training, 101 images for validation and 233 images for testing. The resolution of these images are $960\times720$.

\paragraph{Evaluation Metrics.} For accuracy evaluation, Mean of class-wise Intersection over Union(mIoU) is adopted as the evaluation metric. For speed evaluation, Frames Per Second(FPS) is adopted as the metric.

\begin{figure*}[htbp]
\centering
\subfigure{
\includegraphics[width=0.23\textwidth]{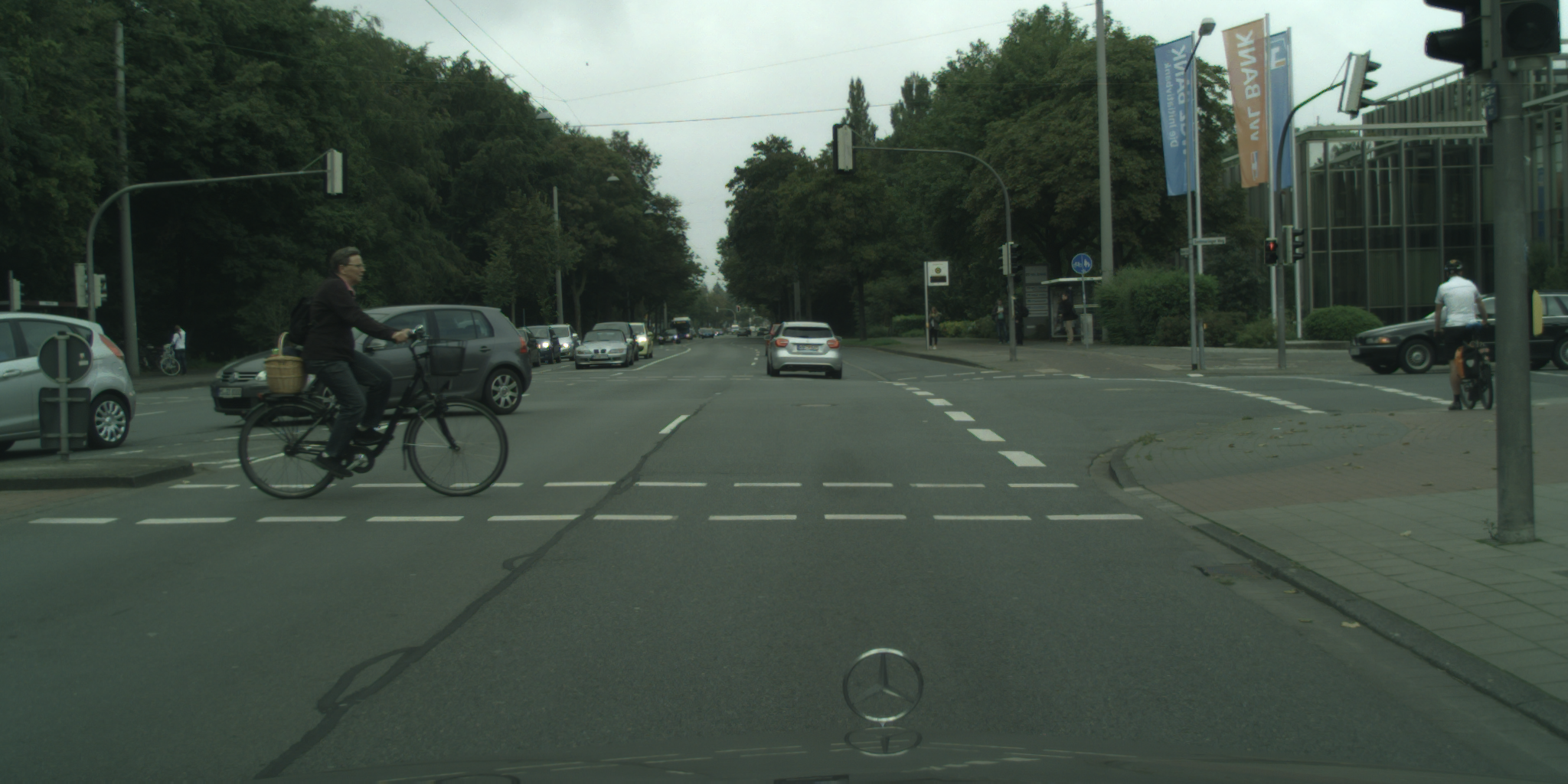}}
\subfigure{
\includegraphics[width=0.23\textwidth]{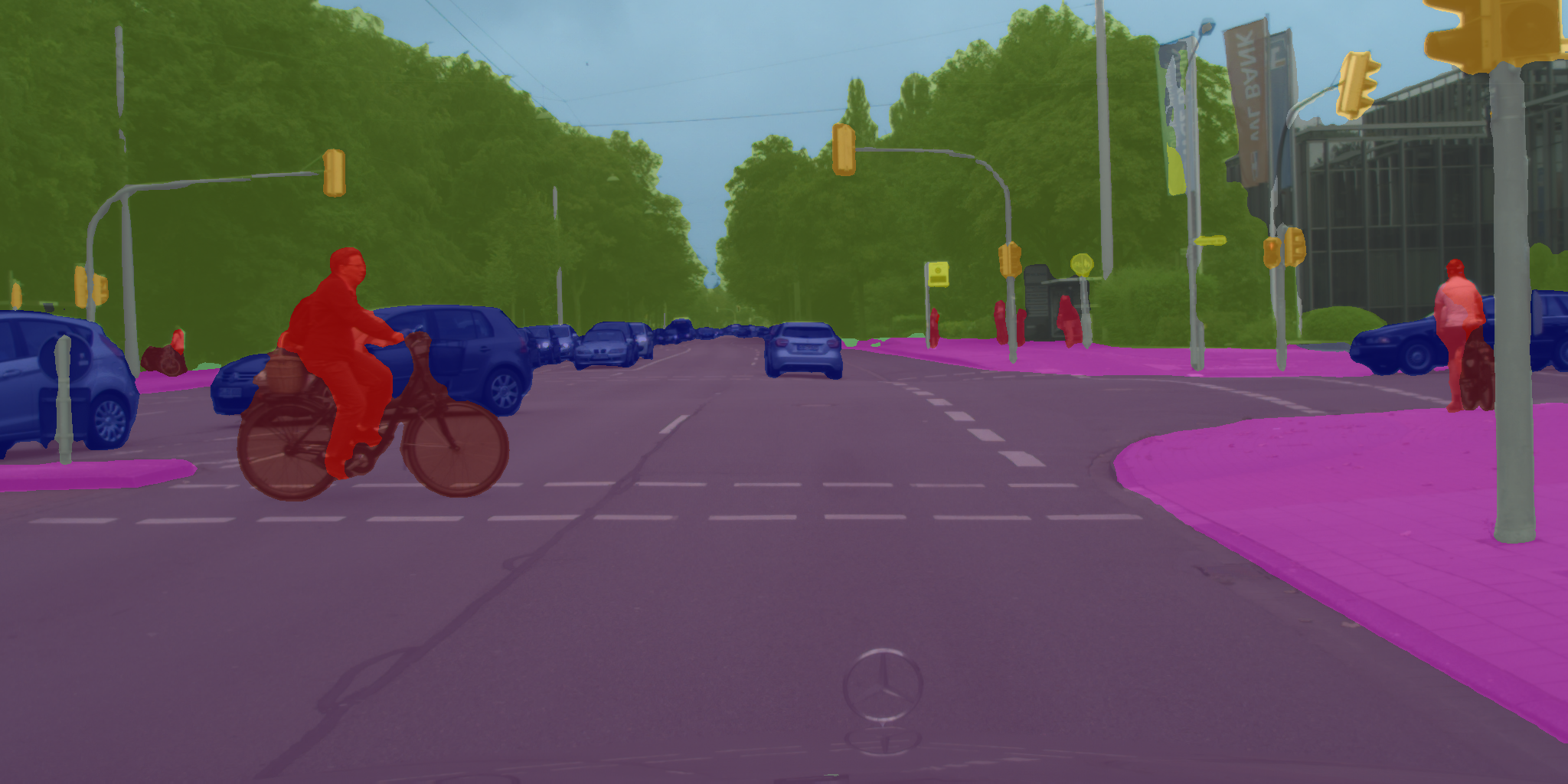}}
\subfigure{
\includegraphics[width=0.23\textwidth]{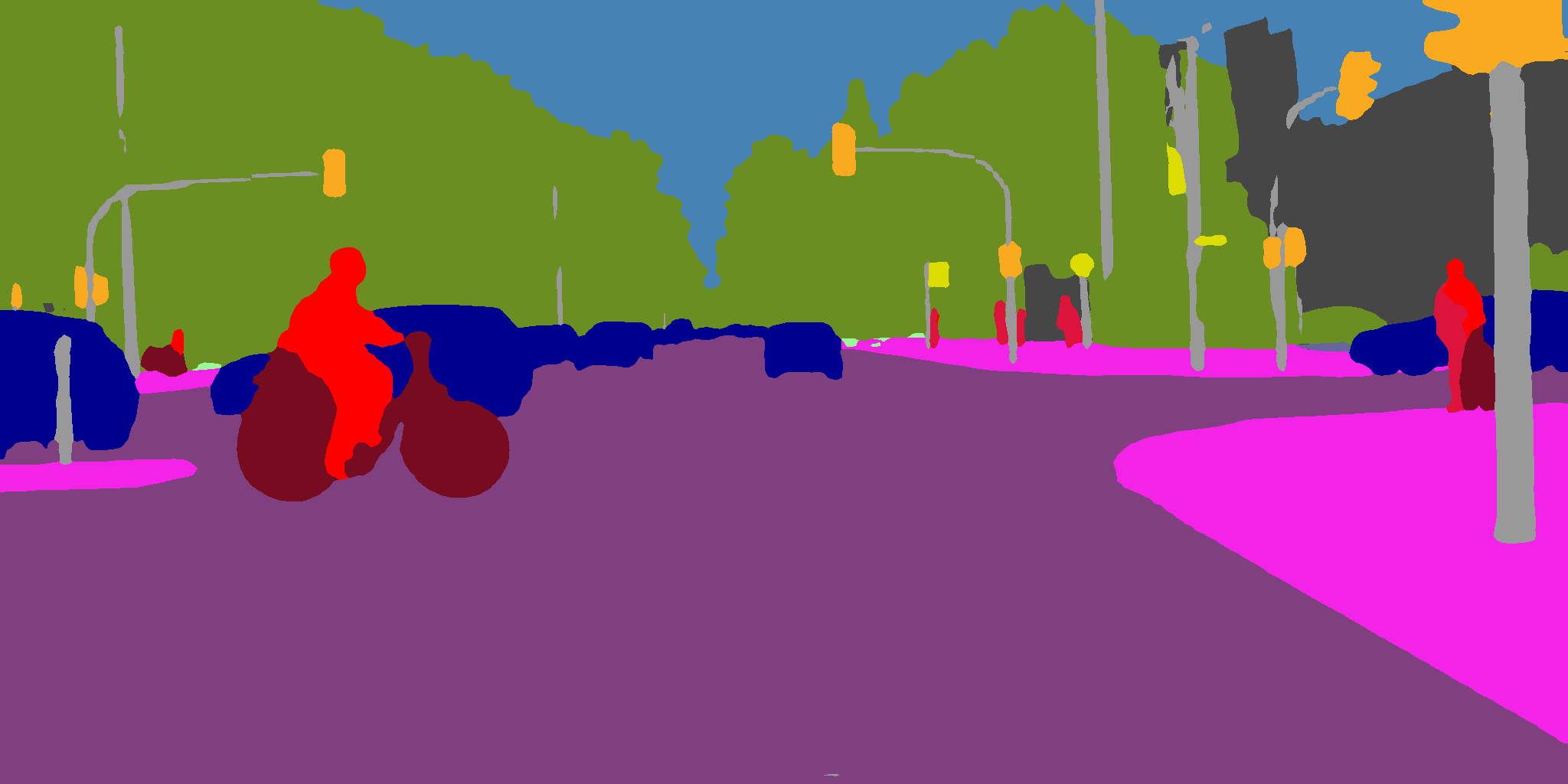}}
\subfigure{
\includegraphics[width=0.23\textwidth]{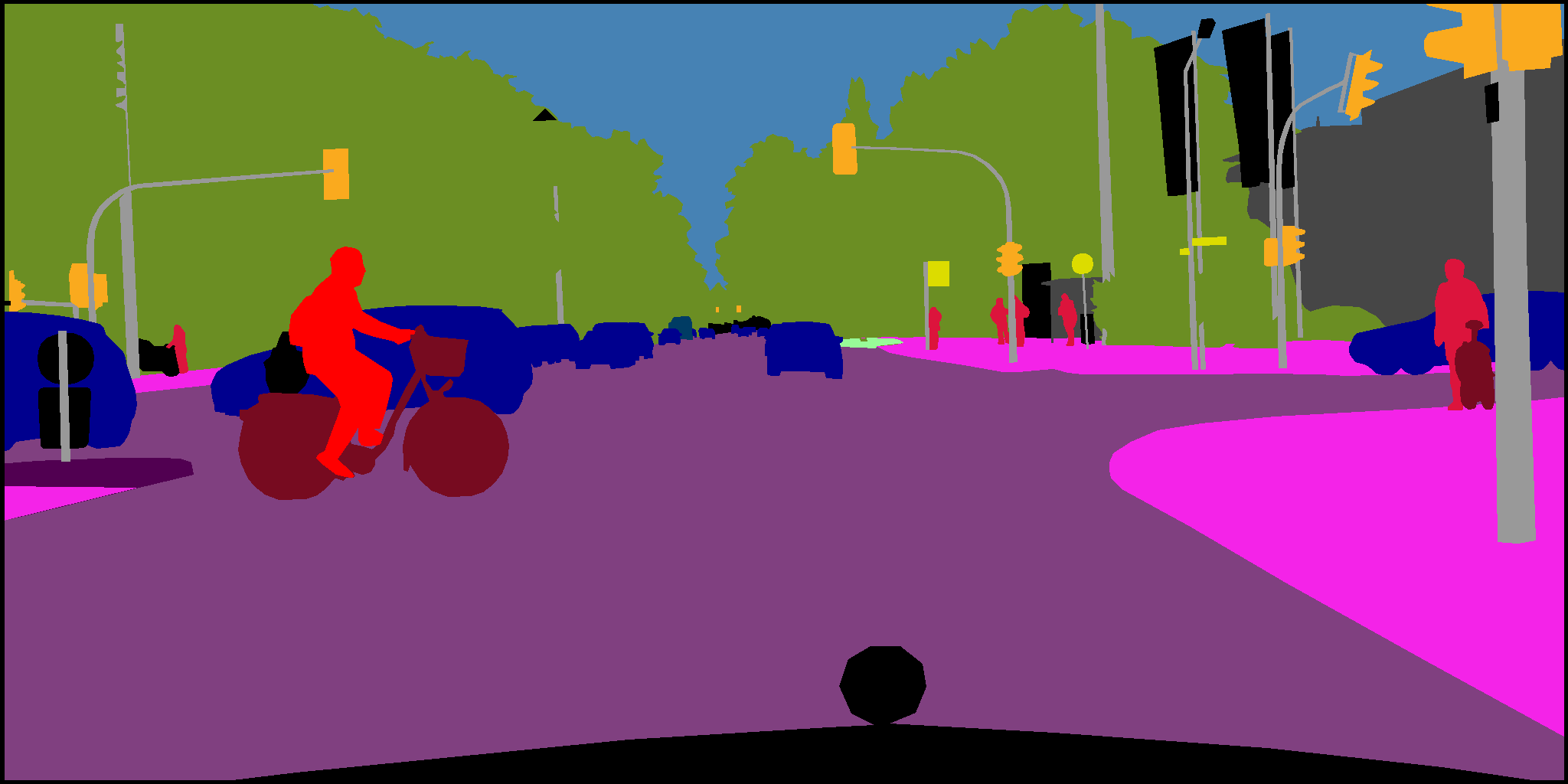}}

\vspace{-1.8mm}

\subfigure{
\includegraphics[width=0.23\textwidth]{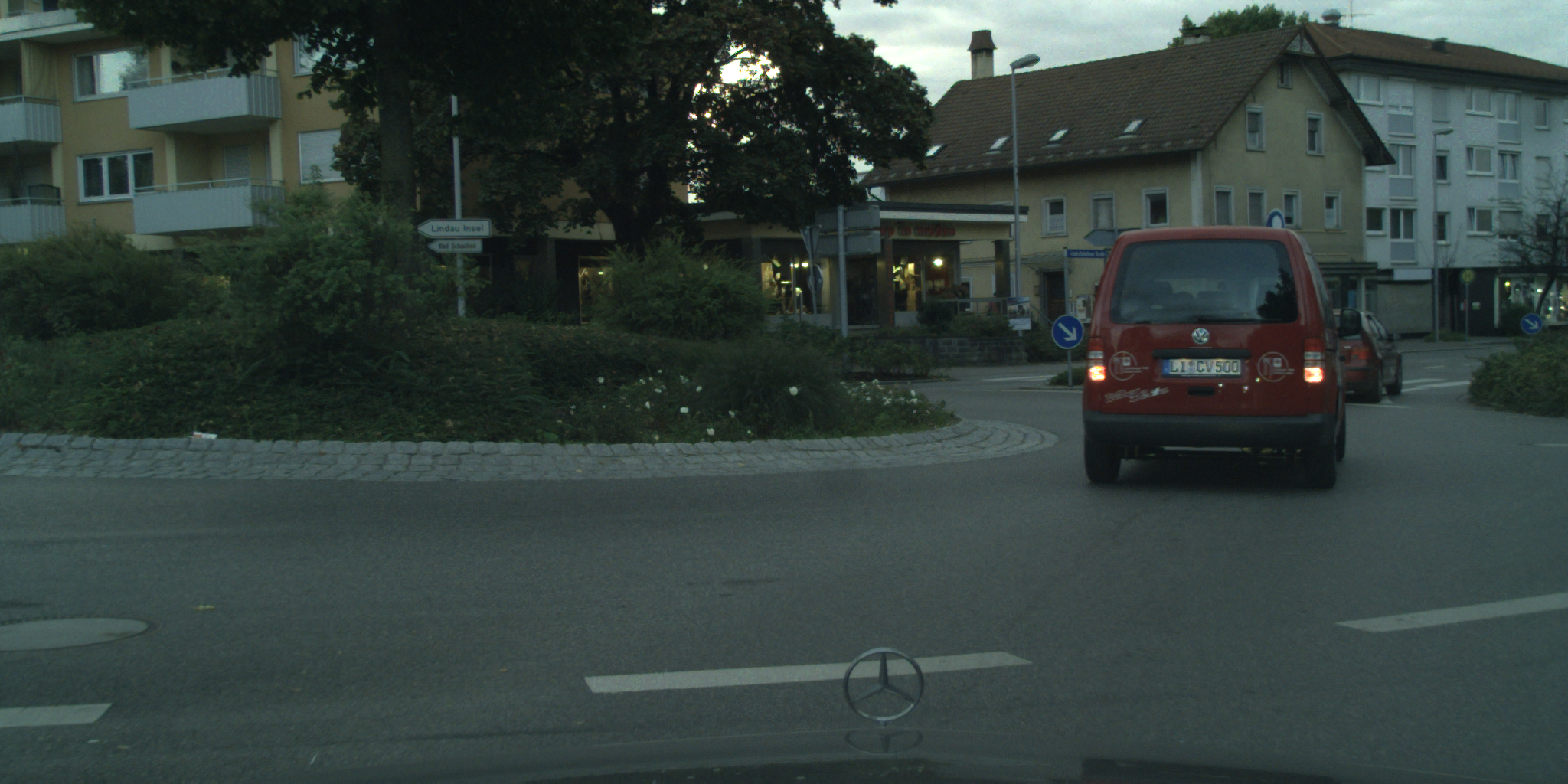}}
\subfigure{
\includegraphics[width=0.23\textwidth]{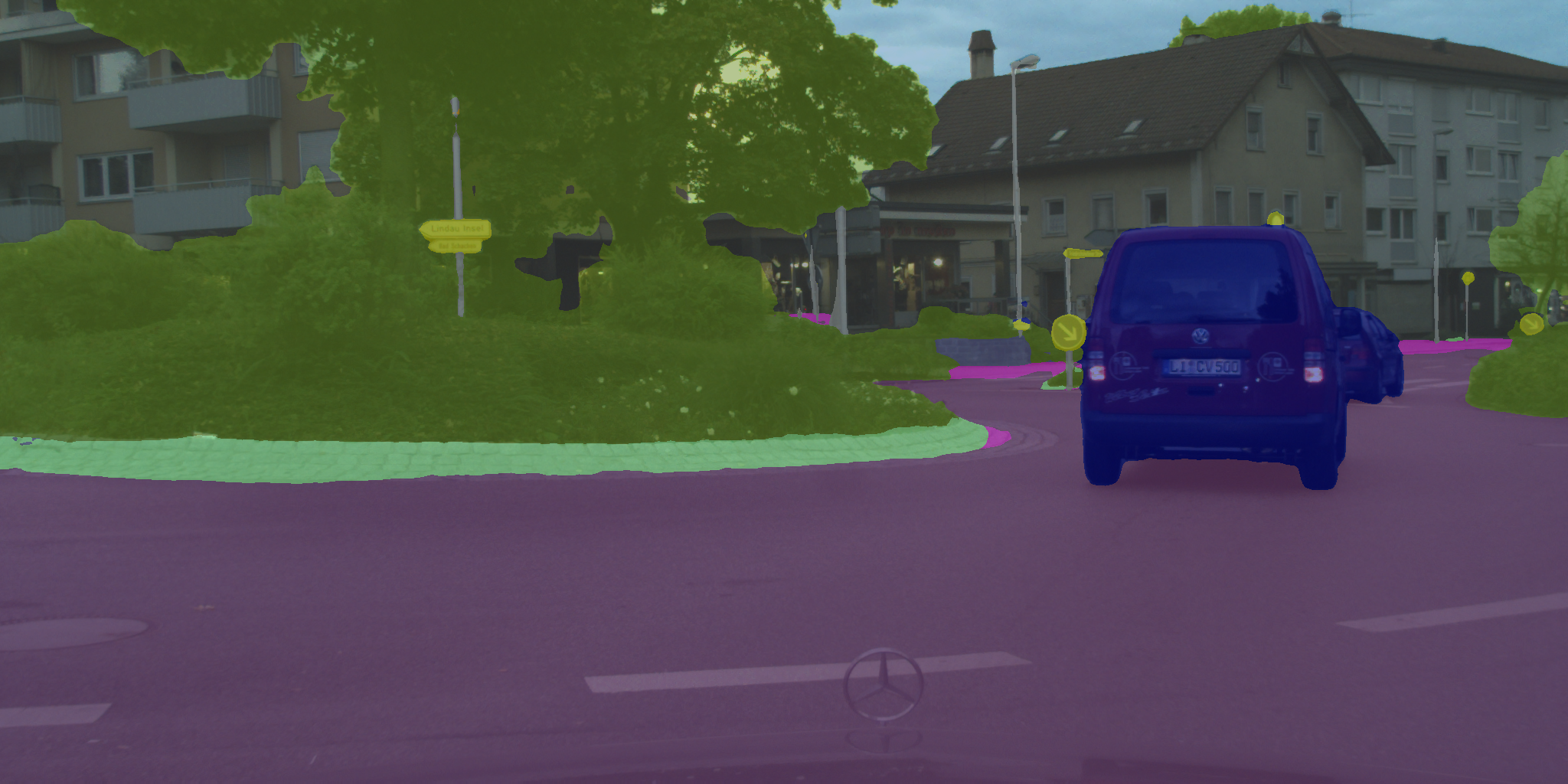}}
\subfigure{
\includegraphics[width=0.23\textwidth]{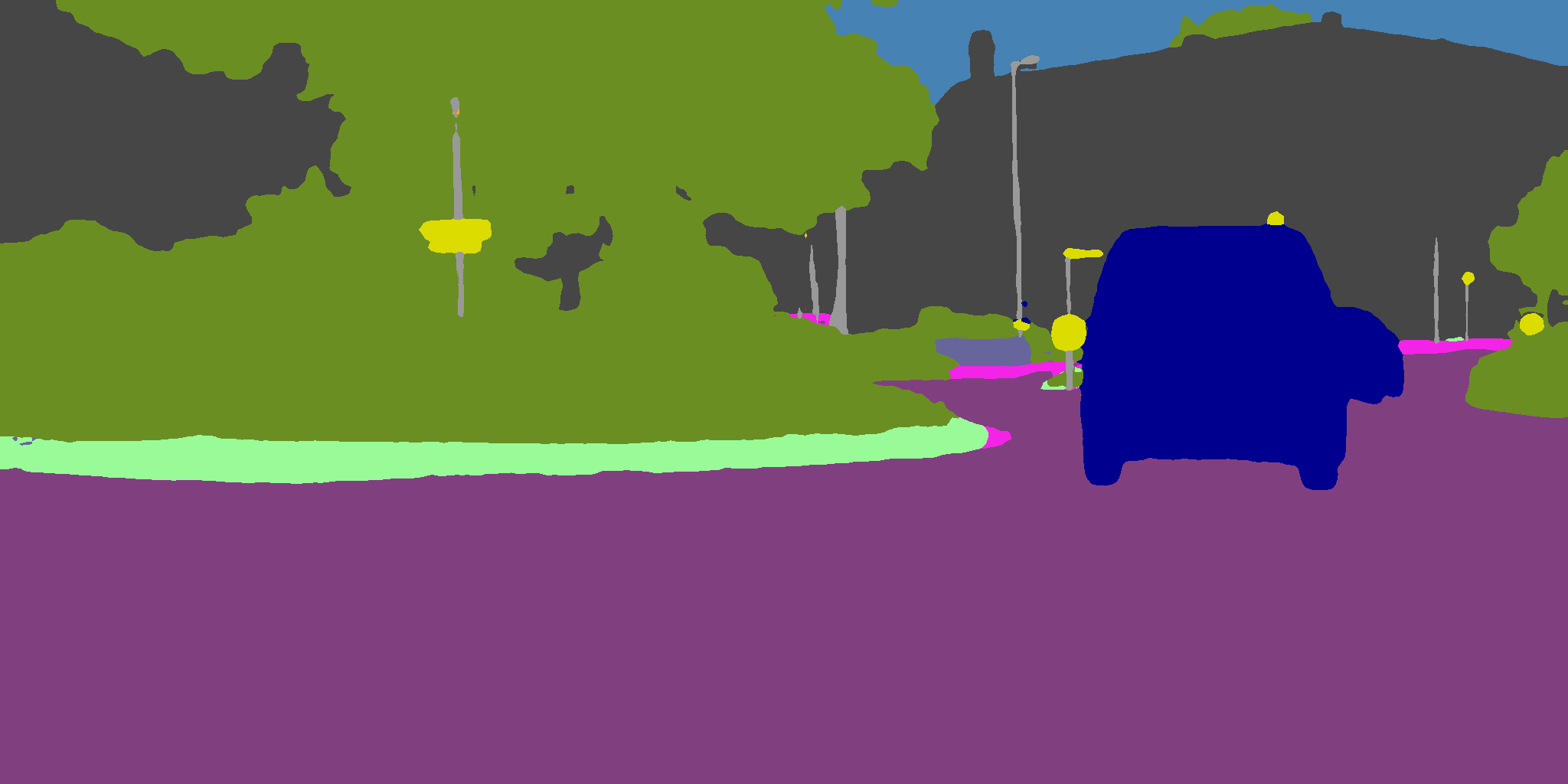}}
\subfigure{
\includegraphics[width=0.23\textwidth]{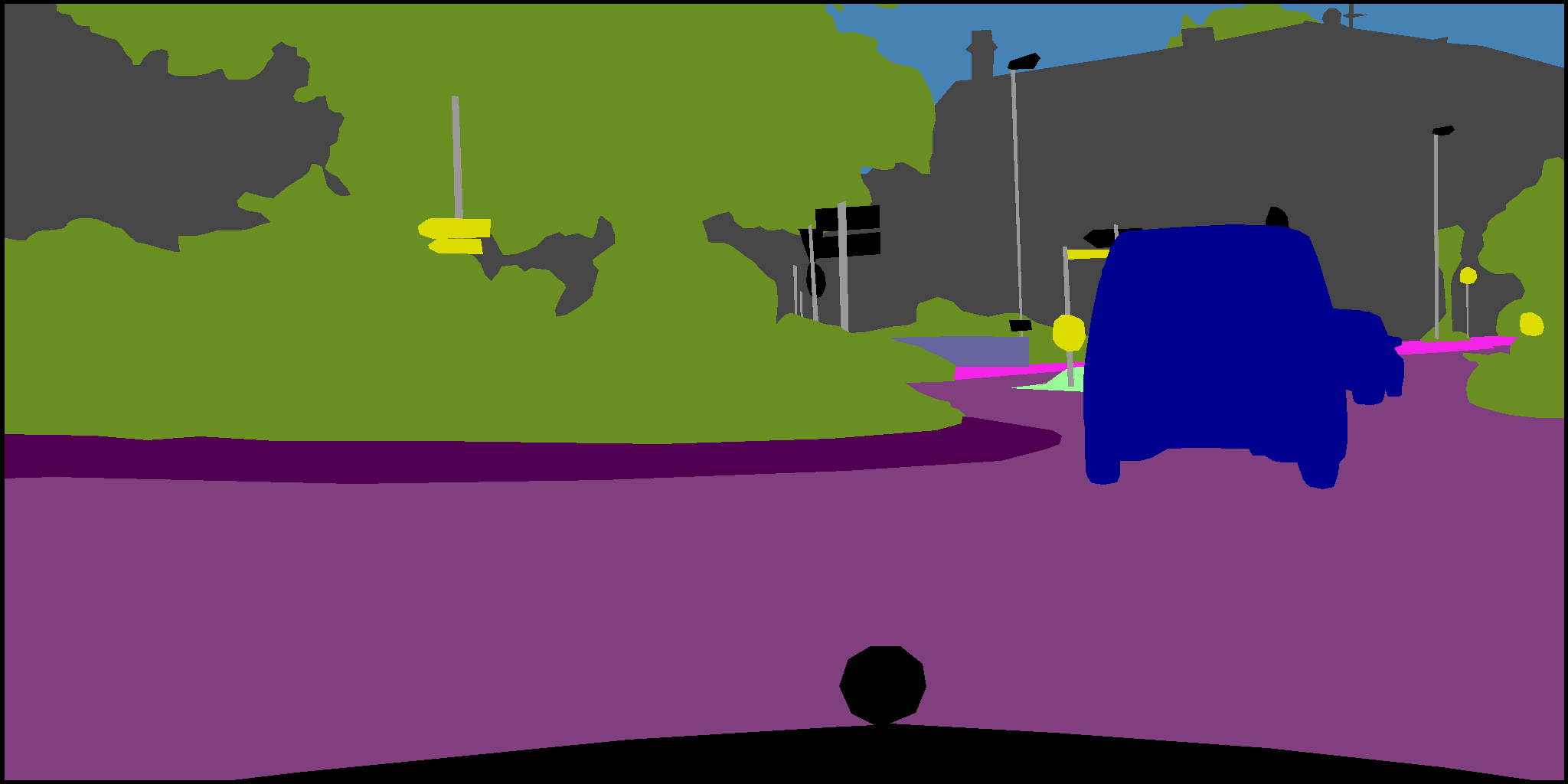}}

\vspace{-1.8mm}

\subfigure{
\includegraphics[width=0.23\textwidth]{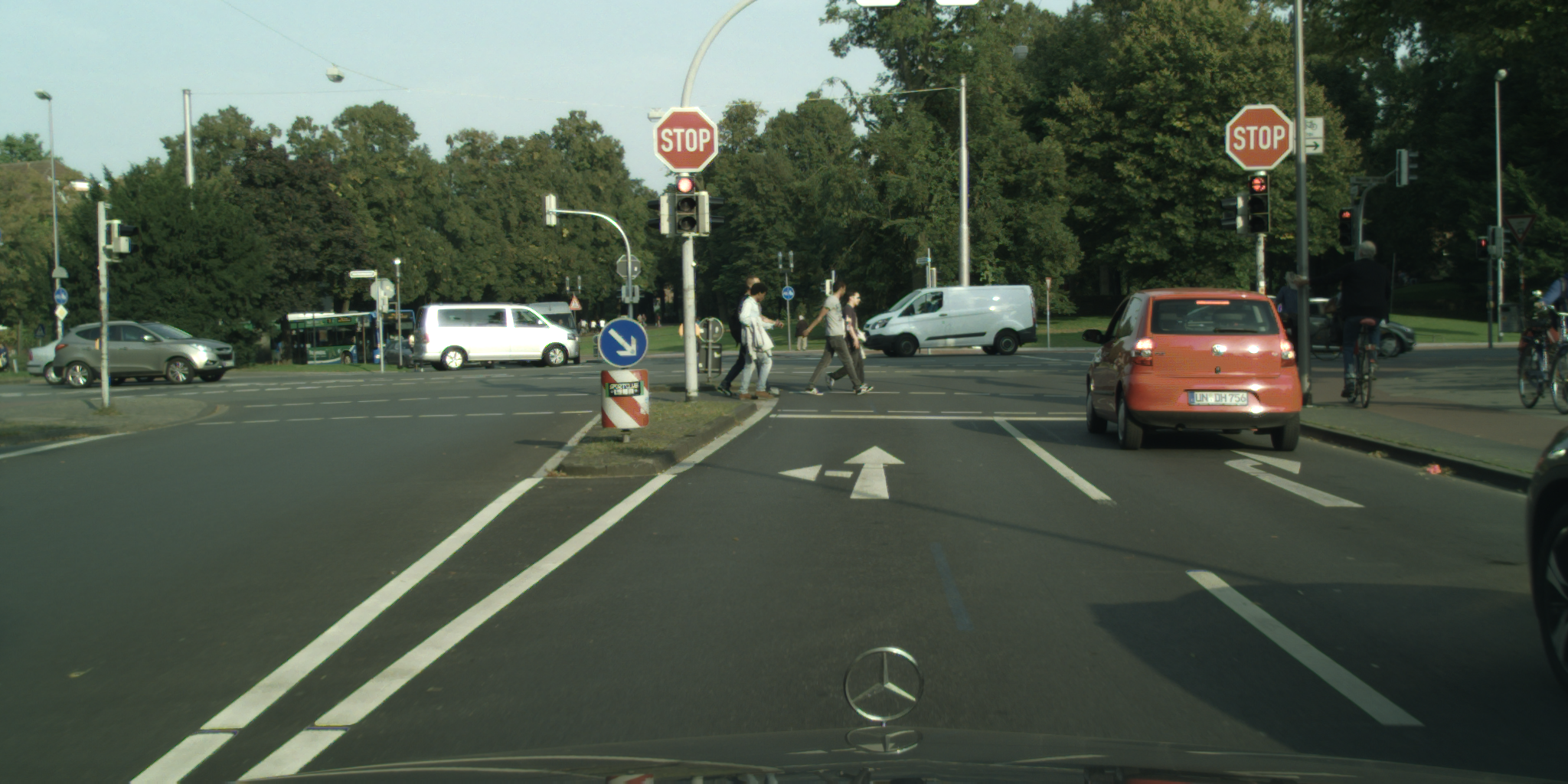}}
\subfigure{
\includegraphics[width=0.23\textwidth]{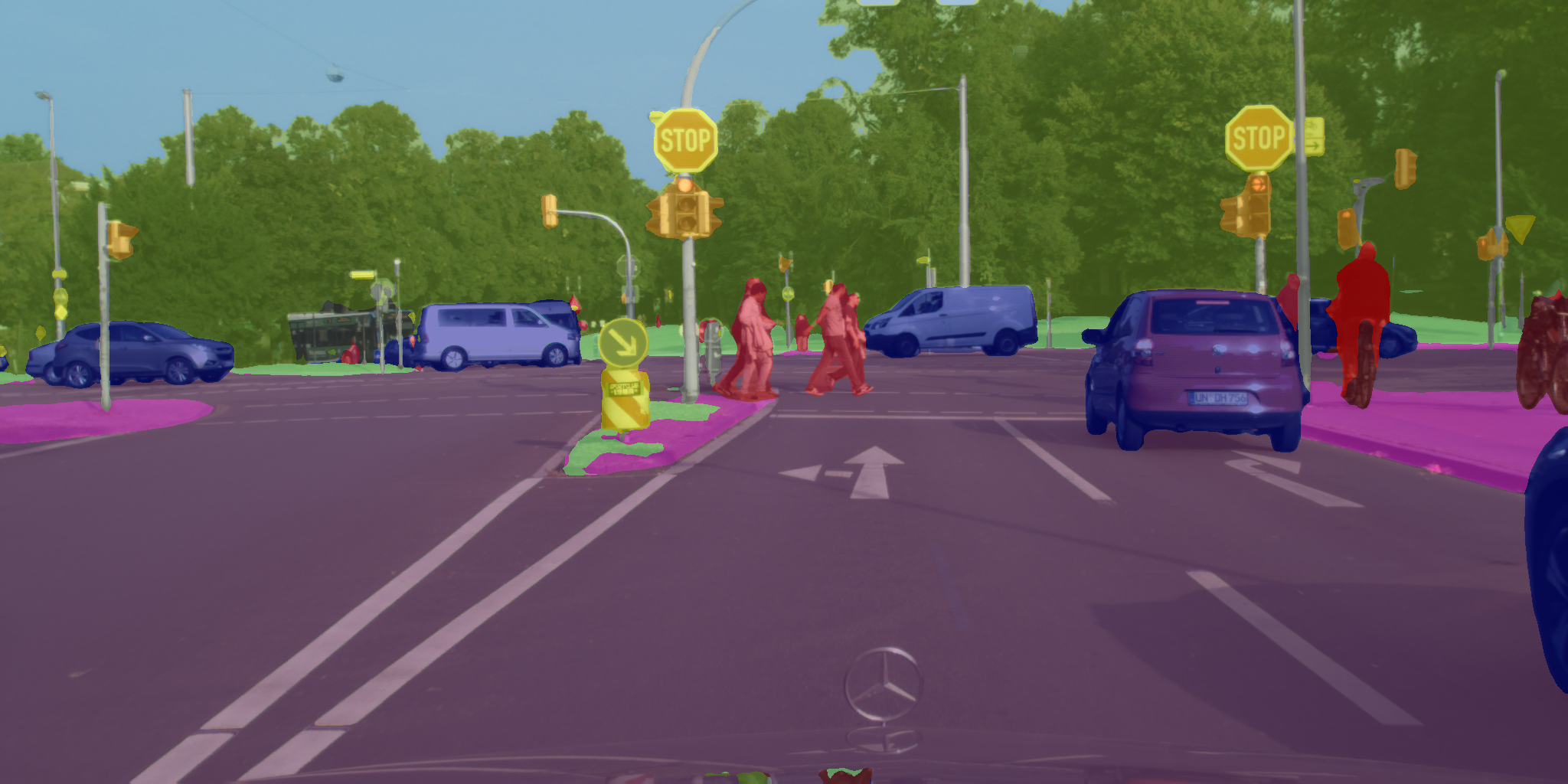}}
\subfigure{
\includegraphics[width=0.23\textwidth]{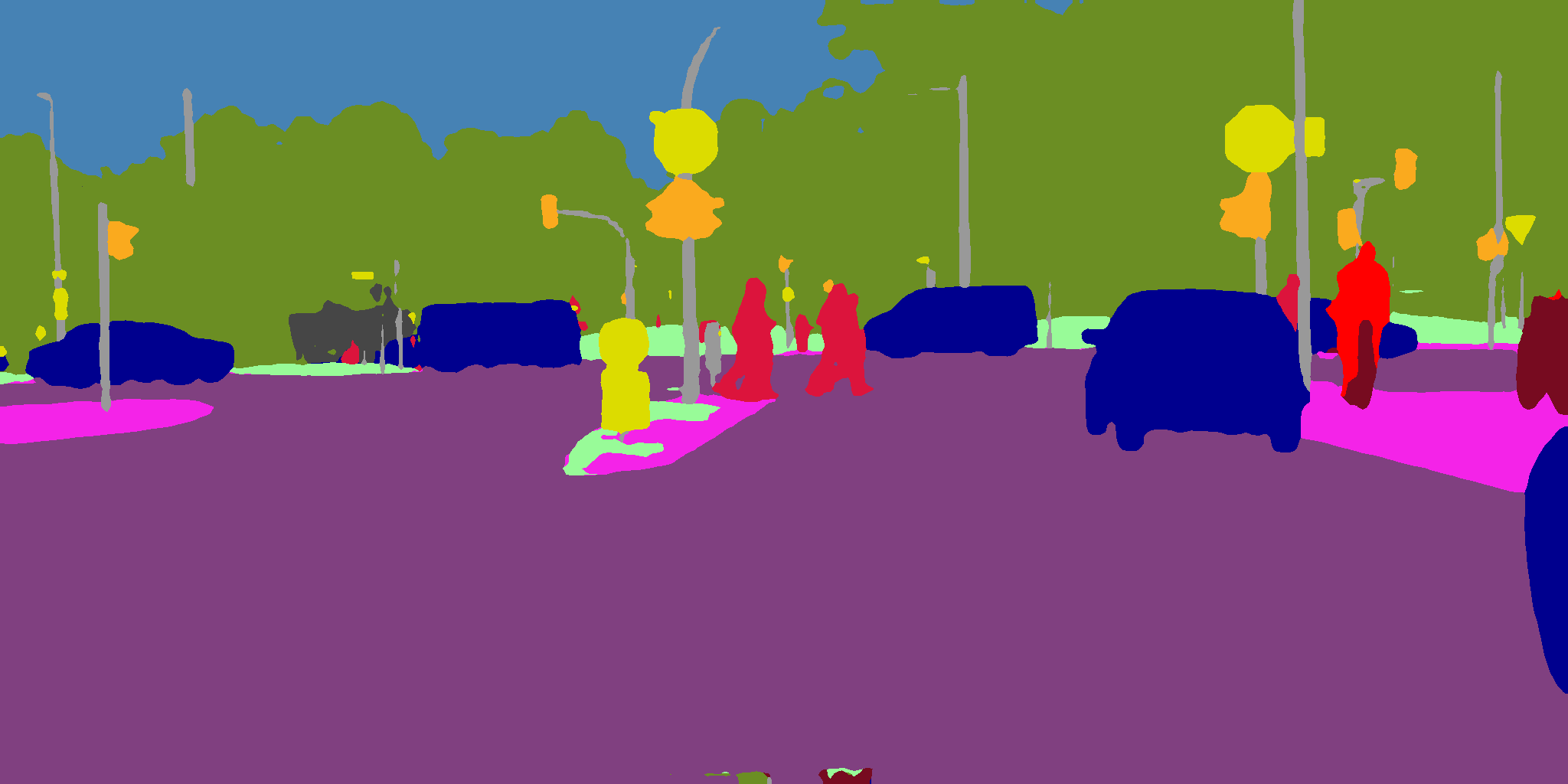}}
\subfigure{
\includegraphics[width=0.23\textwidth]{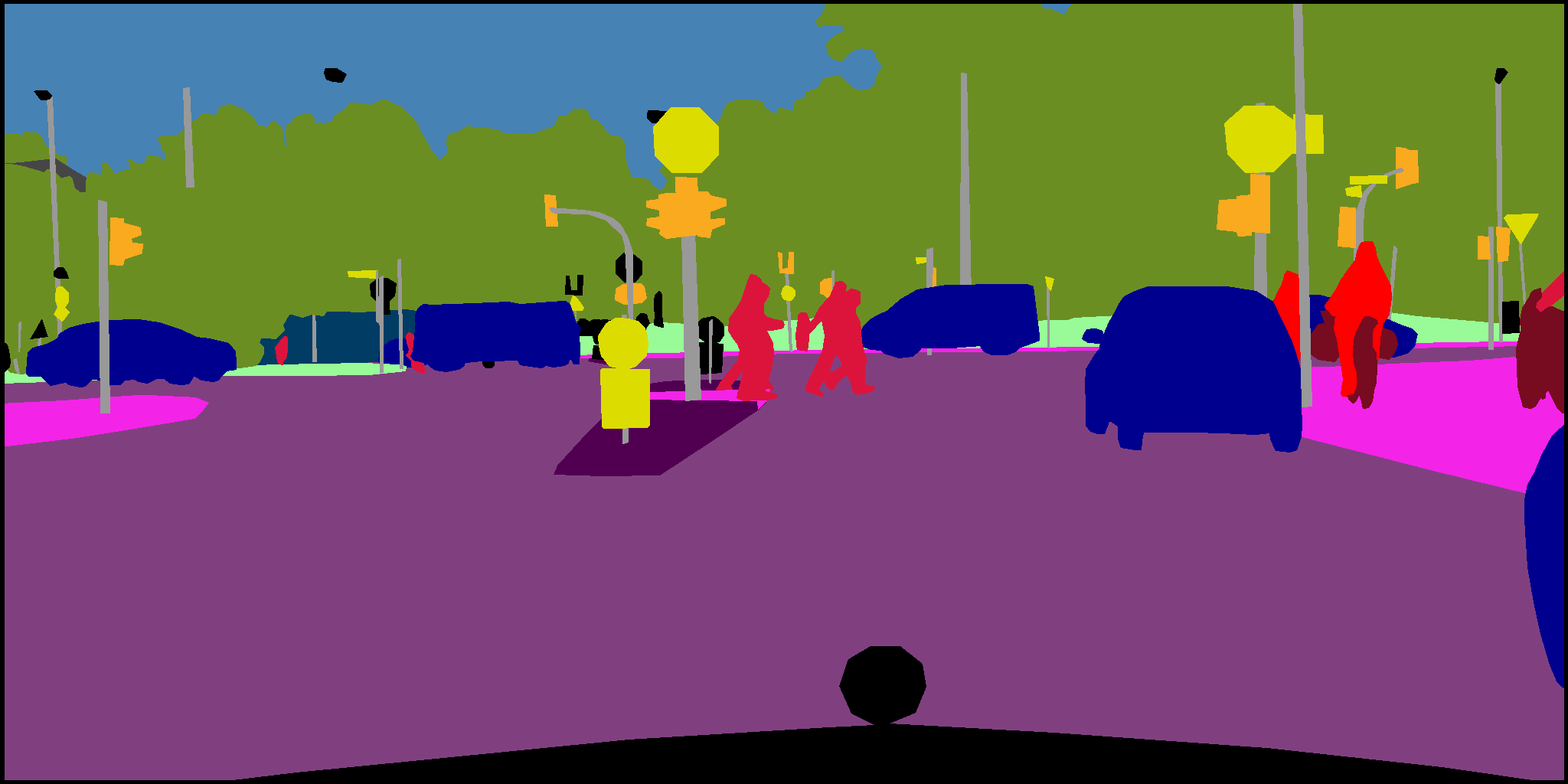}}

\vspace{-1.8mm}

\setcounter{subfigure}{0}
\subfigure[]{
\includegraphics[width=0.23\textwidth]{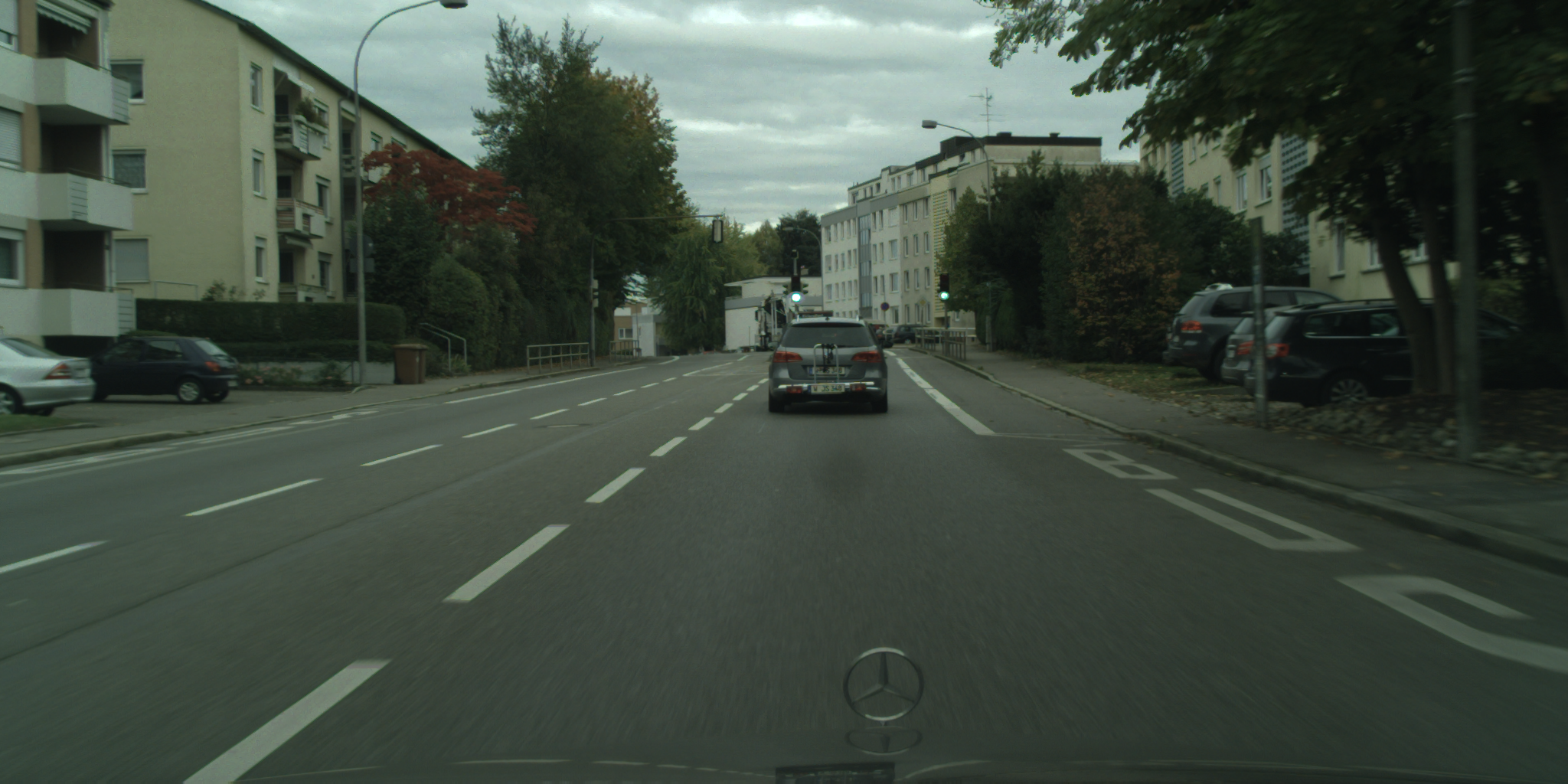}}
\subfigure[]{
\includegraphics[width=0.23\textwidth]{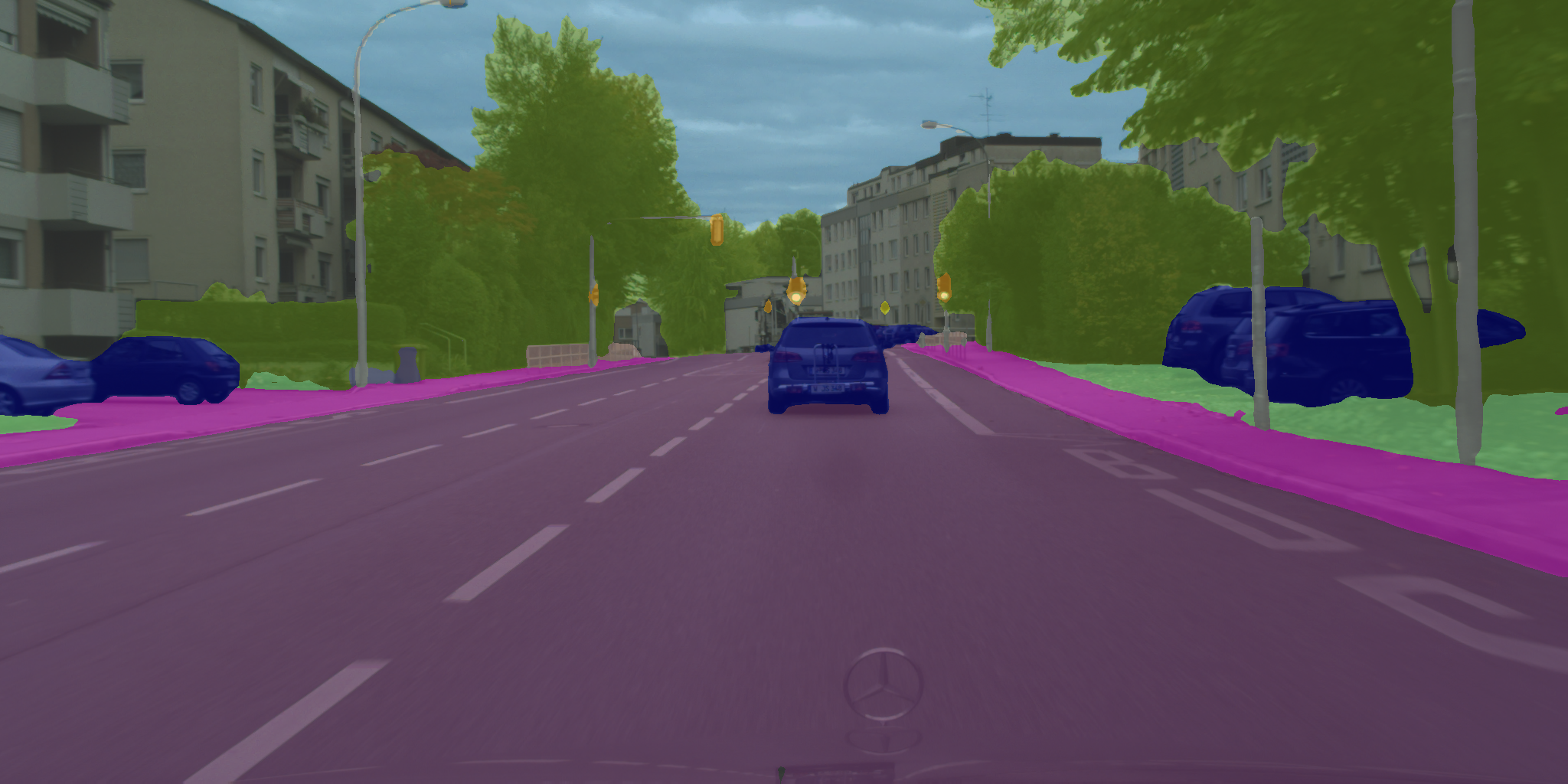}}
\subfigure[]{
\includegraphics[width=0.23\textwidth]{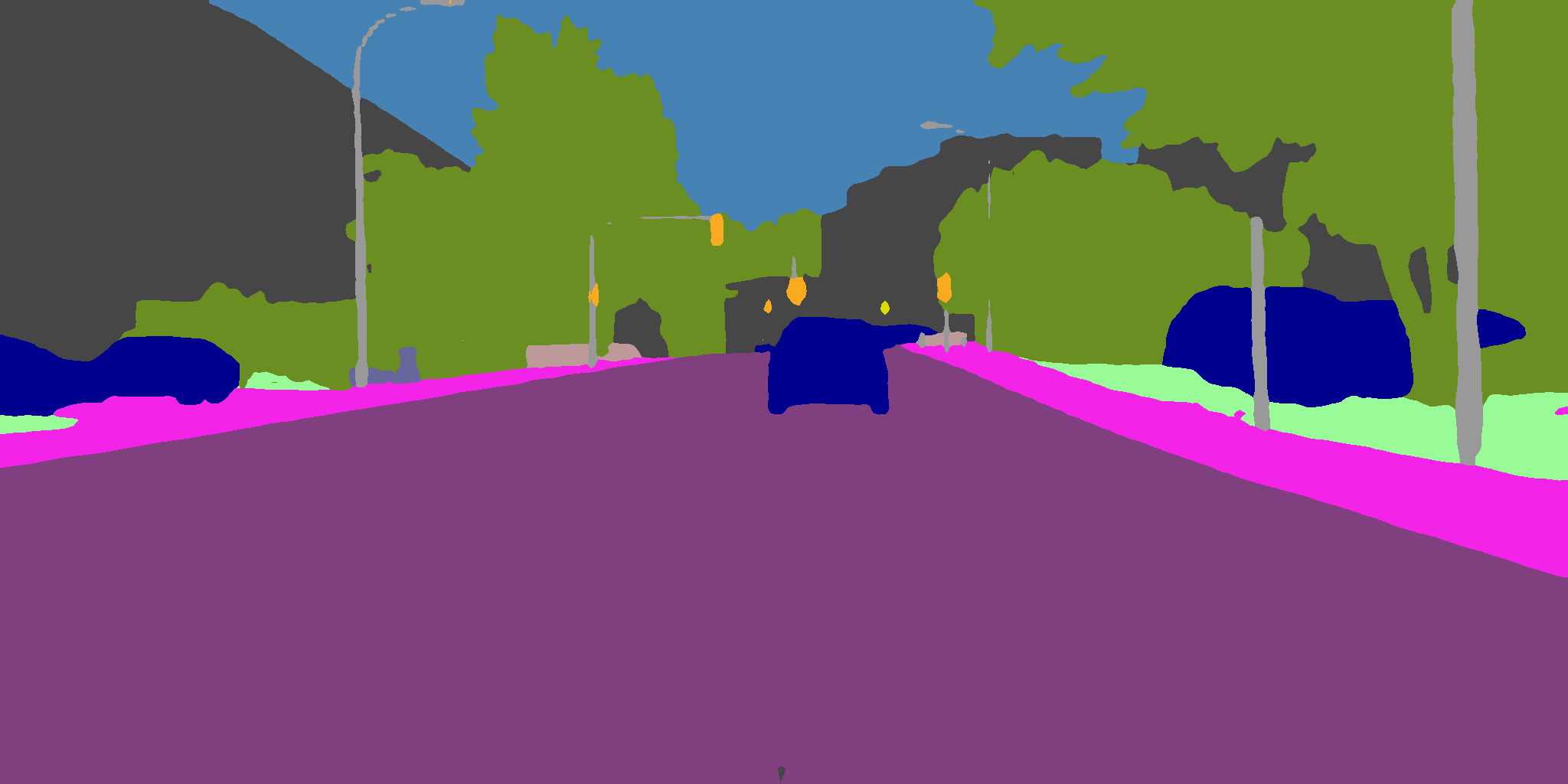}}
\subfigure[]{
\includegraphics[width=0.23\textwidth]{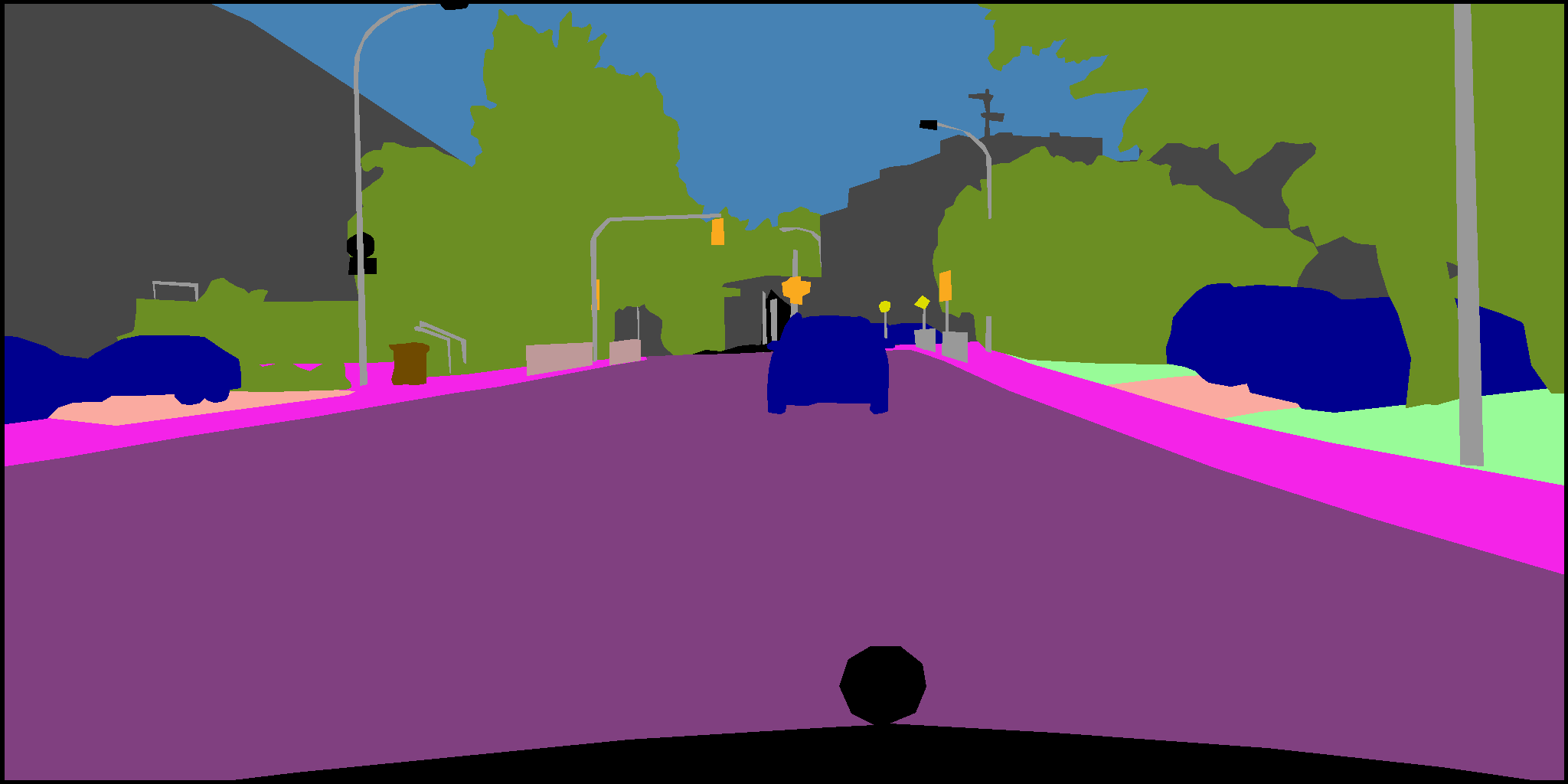}}
    \caption{Visual segmentation results on Cityscapes validation set. The four columns from left to right represent the input image, the output of model(opacity=0.5), the output of model(opacity=1) and the ground truth.}
    \label{fig7_visual segmentation results}
\end{figure*}

\subsection{Implementation Details}
\paragraph{Training Settings.}
As with most experiments, we choose mini-batch stochastic gradient descent(SGD) with a momentum of 0.9 and weight decay of $1\times5e^{-4}$ as the optimizer. The batch sizes of the Cityscapes and CamVid datasets are set to 16 and 24, respectively. Following the common configuration, 'poly' learning rate policy is adopted, where the initial rate is multiplied by $(1-\frac{iter}{max\_iter})^{power}$. We set the power as 0.9, the initial learning rate as 0.01 and the minimum learning rate as $1\times e^{-4}$. We utilize the Cityscapes dataset to train the model for 160,000 iterations, utilize CamVid dateset to train the model for 10000 iterations. 
For data augmentation, random-crop, random-resize, and random-flip are included. For training Cityscapes, the cropped resolution is $512\times1024$ and the scale ranges in [0.125,0.5]. For training CamVid, the cropped resolution is $960\times720$ and the scale ranges in [0.5,2.5]. We conduct all experiments based on Pytorch-1.11, CUDA-11.3, CUDNN-8.4.1 environments. And all experiments are performed using MMsegmentation\cite{mmseg2020} on two GTX-1080Ti GPUs .
\paragraph{Inference Settings.}
In the inference phase, we export Cross-CBAM to ONNX format and utilize TensorRT for model inference. For Cityscapes, the resolution of the input images are $512\times1024$ and $768\times1536$. For CamVid, the resolution of the input images are $960\times720$. All inference experiments are conducted on a single NVIDIA GTX1080Ti GPU with batch size of 1 under CUDA-11.3, CUDNN-8.2.2.1, TensorRT-8.2.3.0.

\begin{figure*}[htbp]
\subfigure[]{
\begin{minipage}{0.23\linewidth}
\includegraphics[width=1\textwidth]{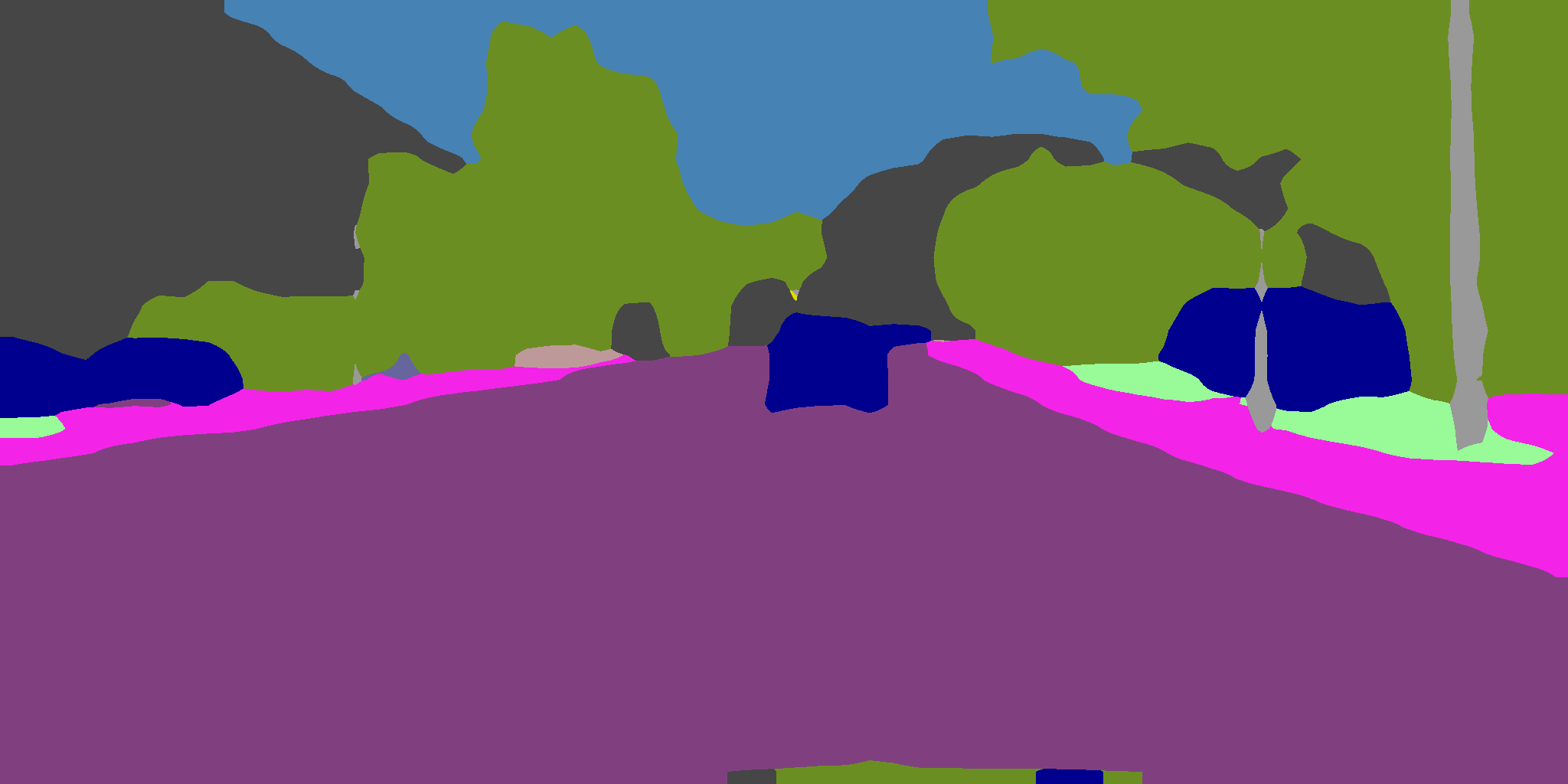}\\
\vspace{-2mm}
\includegraphics[width=1\textwidth]{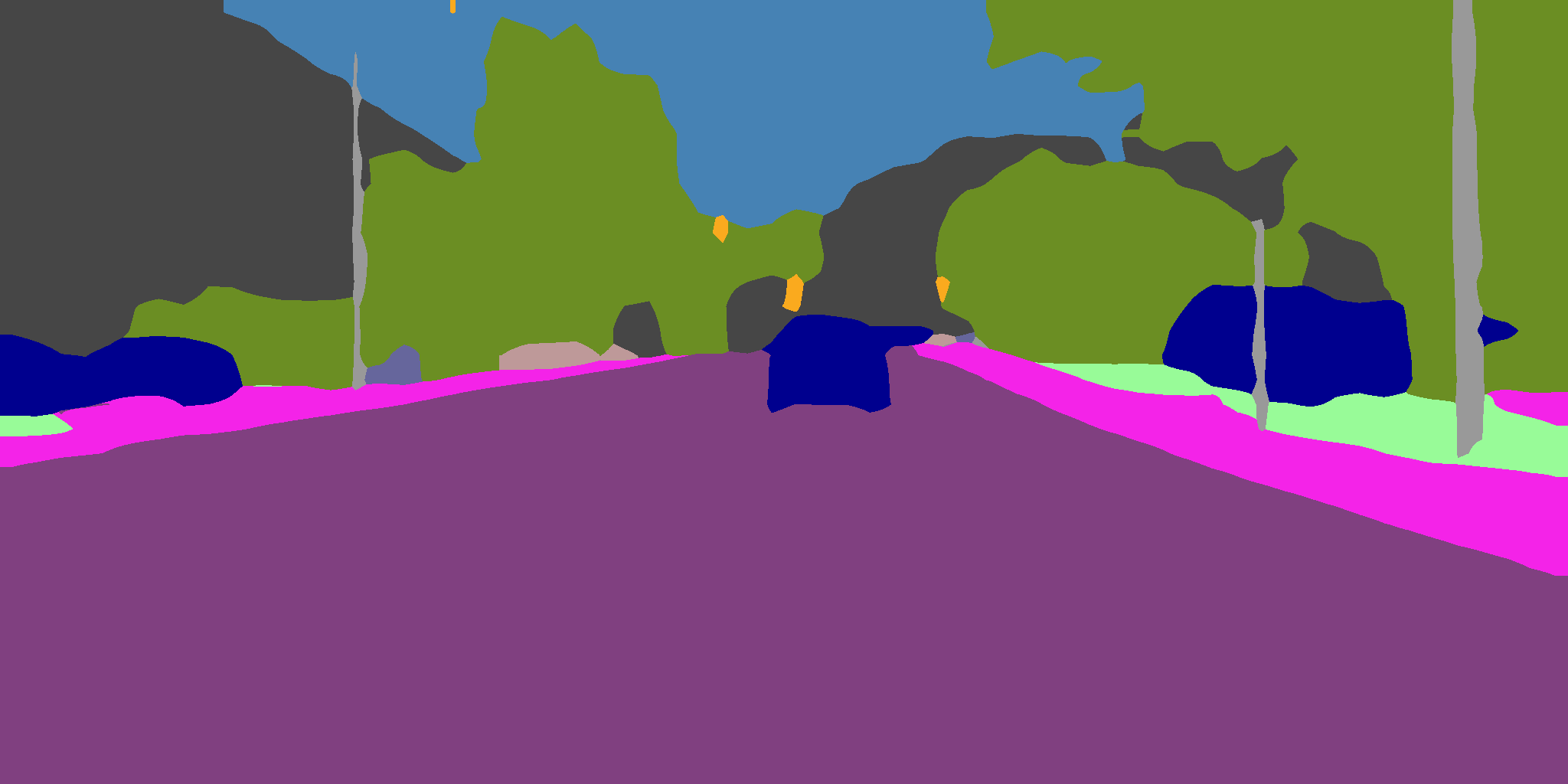}\\
\vspace{-2mm}
\includegraphics[width=1\textwidth]{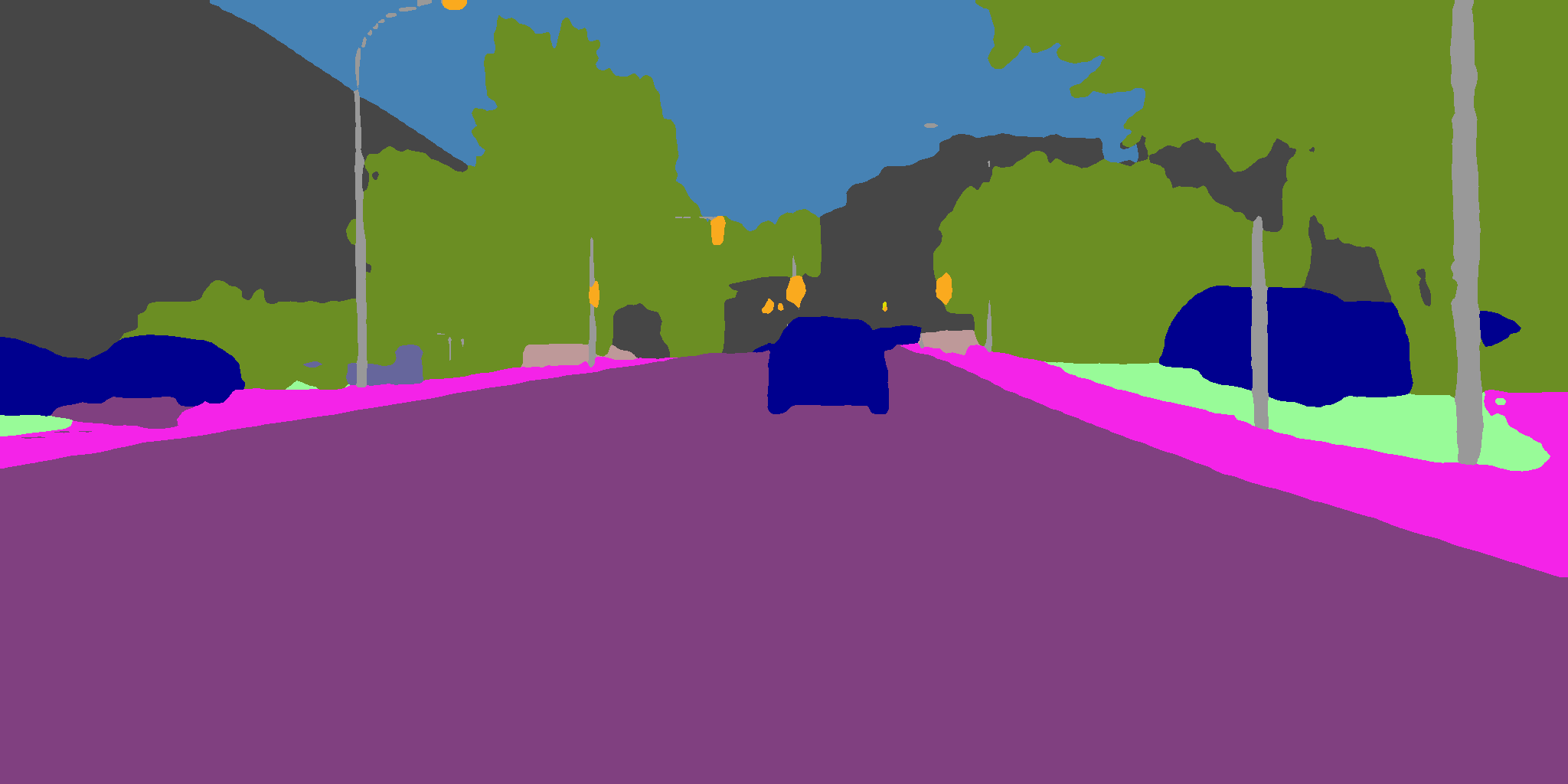}\\
\vspace{-2mm}
\includegraphics[width=1\textwidth]{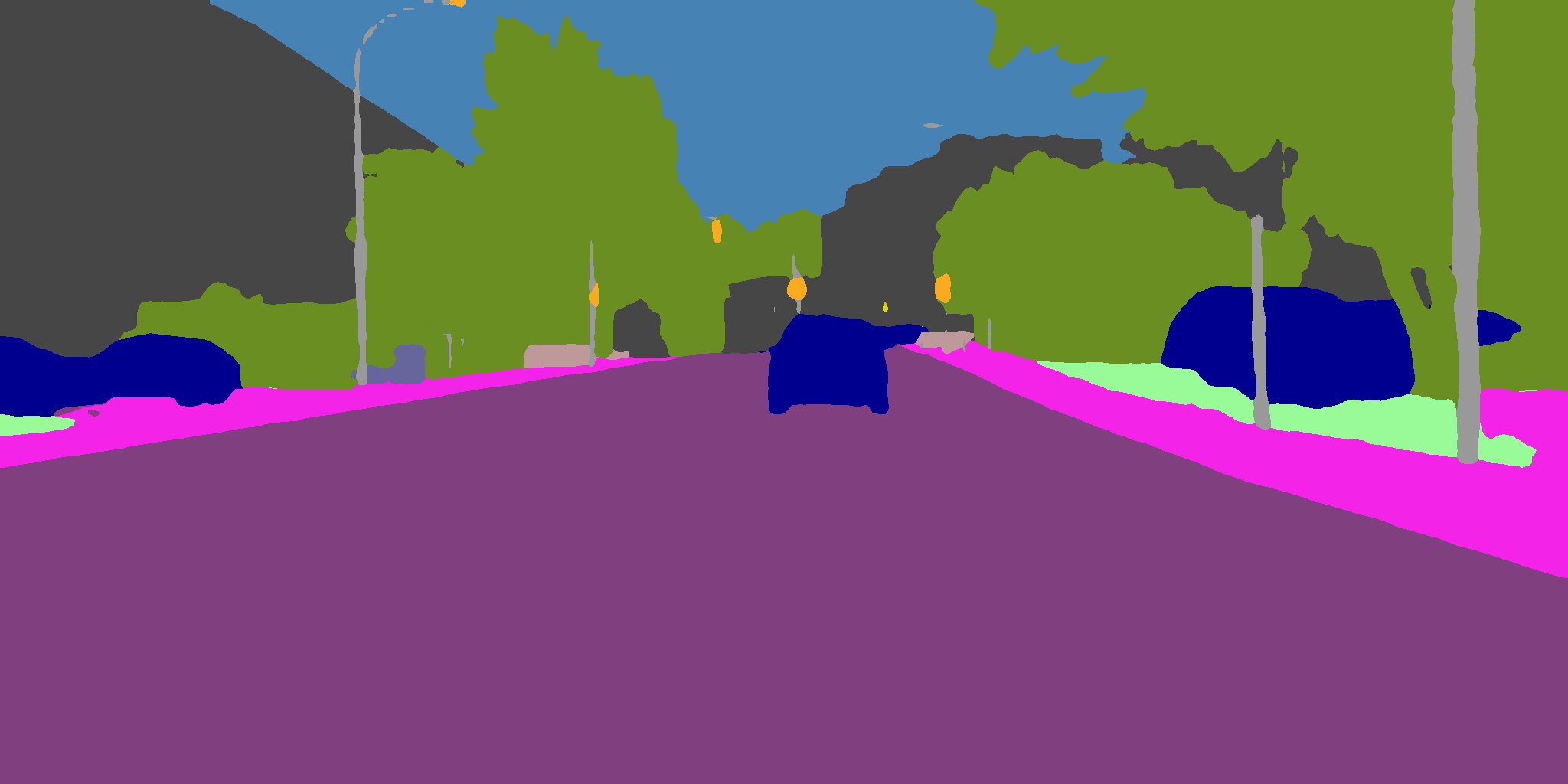}\\
\vspace{-2mm}
\includegraphics[width=1\textwidth]{lindau_000004_000019_STDC-CCBAM-SEASPP.png}\\
\vspace{-2mm}
\includegraphics[width=1\textwidth]{lindau_000004_000019_gtFine_color.png}\\
\end{minipage}
}
\hspace{-2mm}
\subfigure[]{
\begin{minipage}{0.23\linewidth}
\includegraphics[width=1\textwidth]{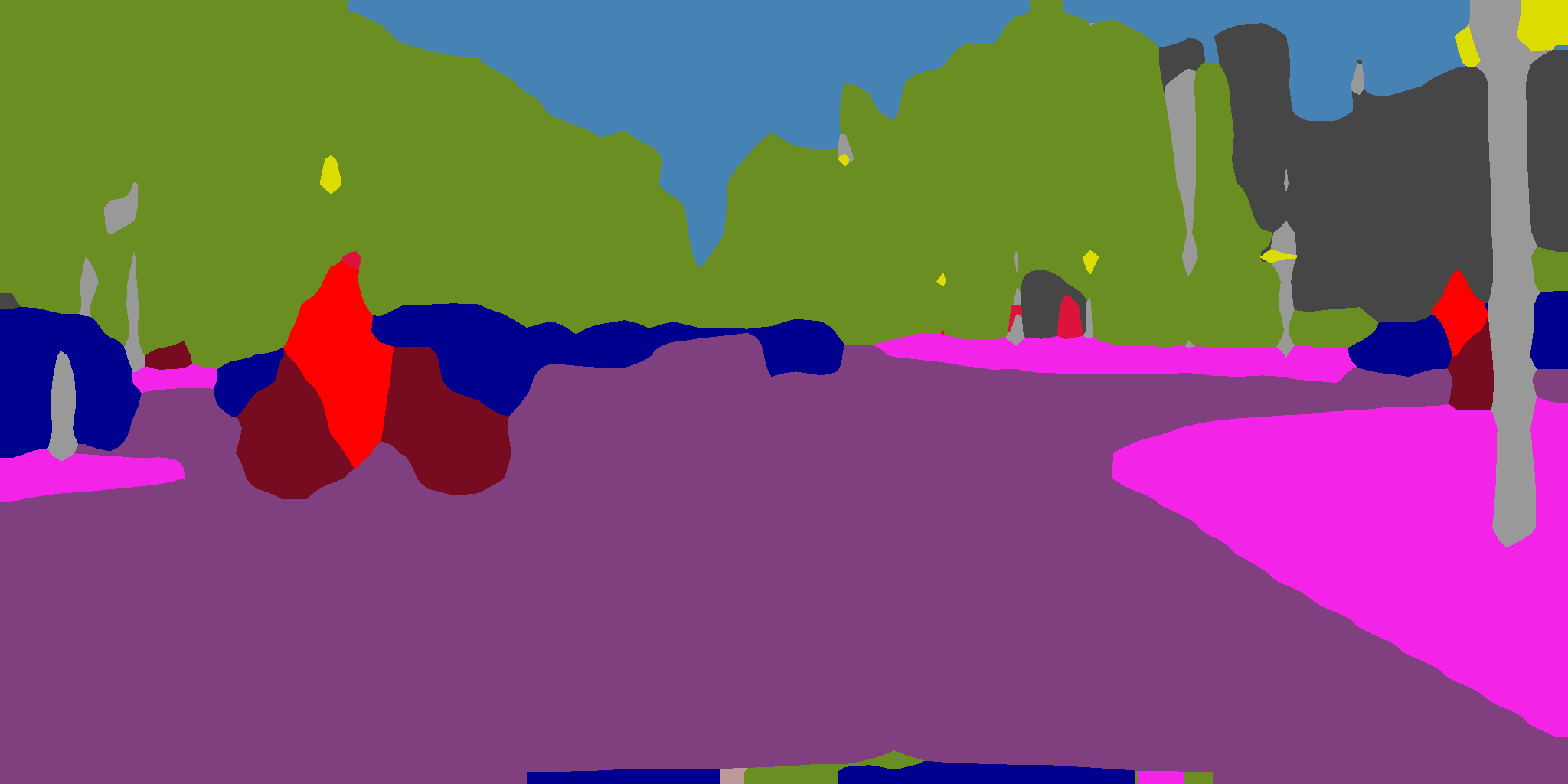}\\
\vspace{-2mm}
\includegraphics[width=1\textwidth]{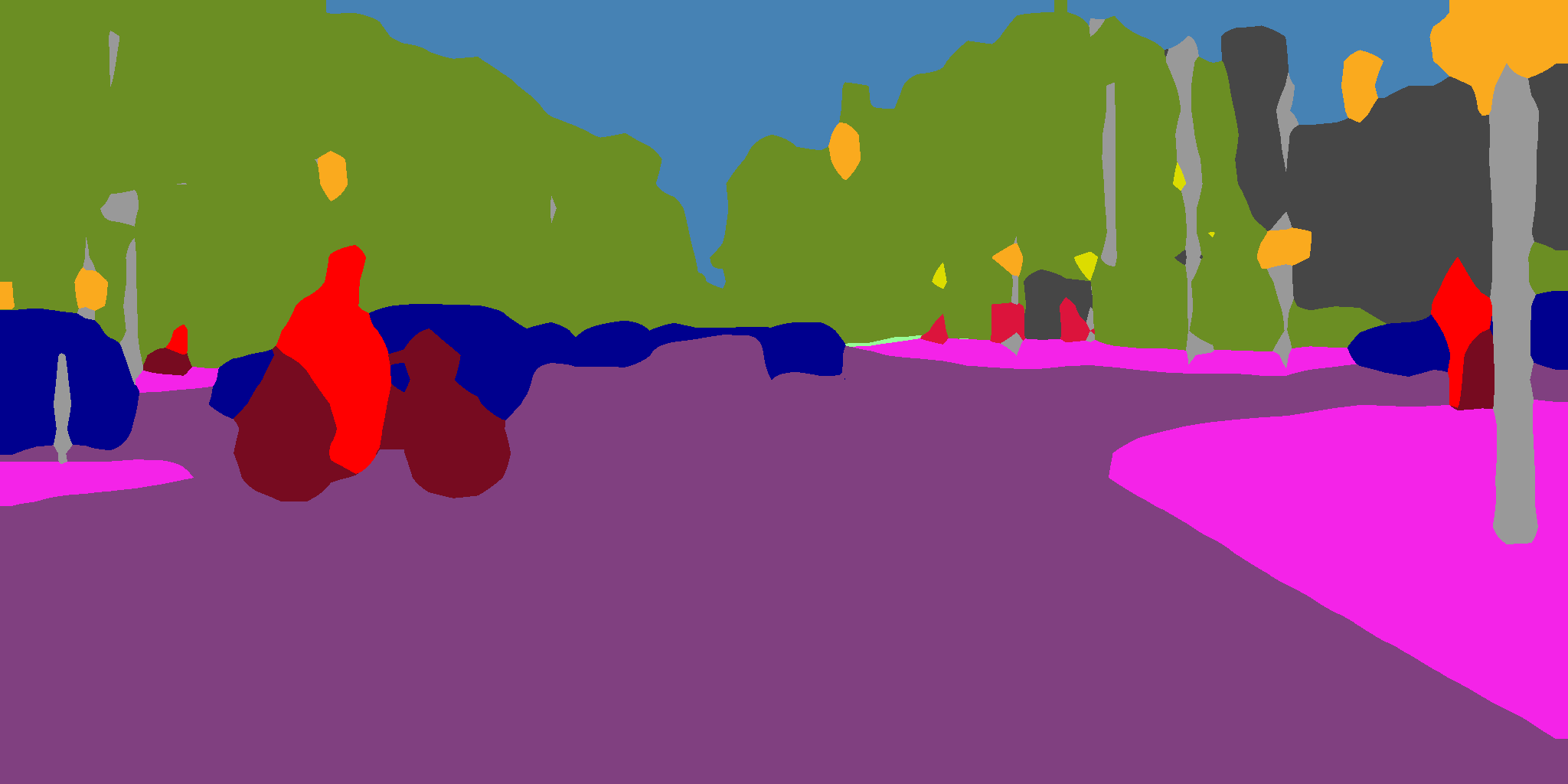}\\
\vspace{-2mm}
\includegraphics[width=1\textwidth]{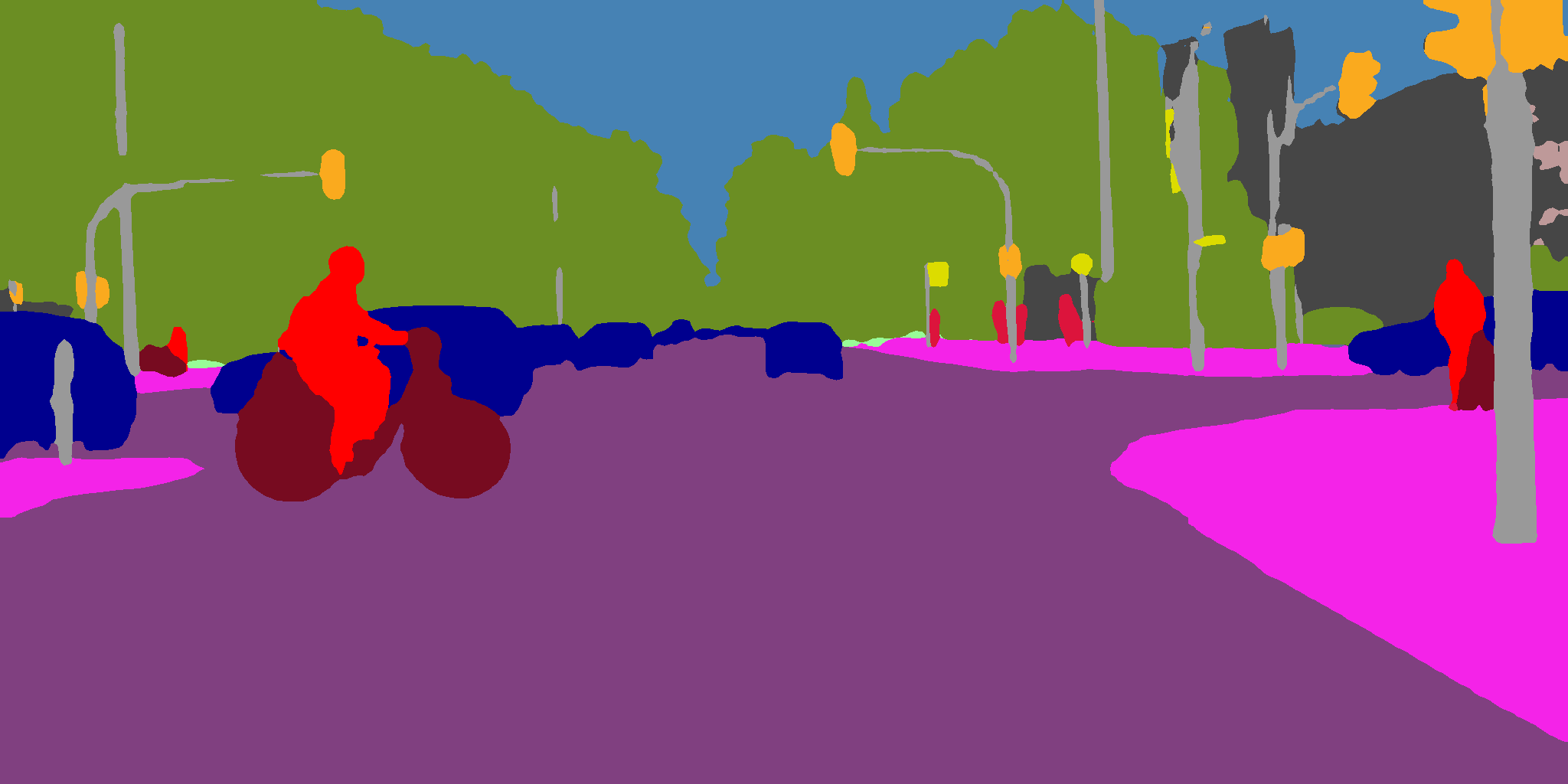}\\
\vspace{-2mm}
\includegraphics[width=1\textwidth]{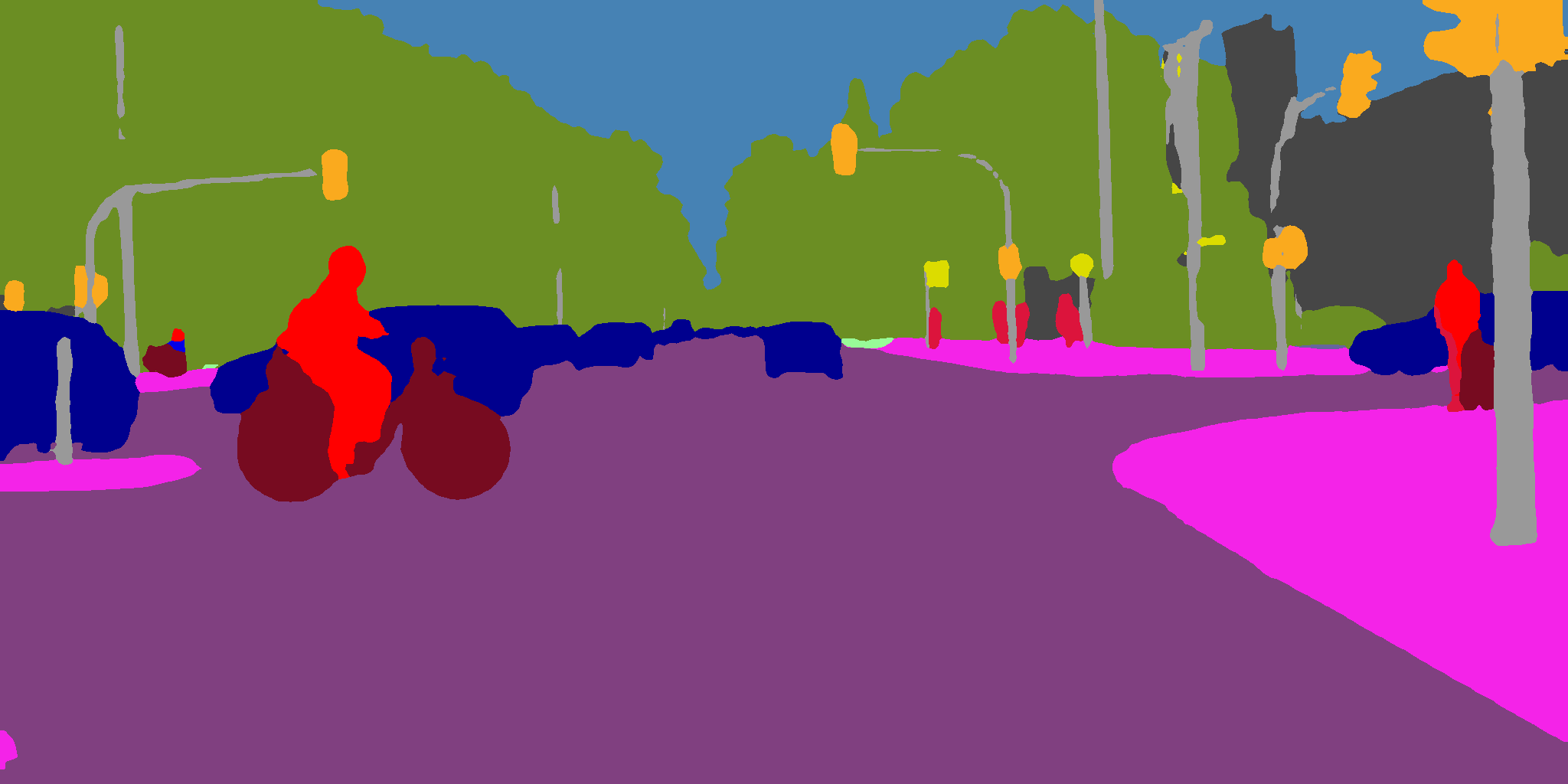}\\
\vspace{-2mm}
\includegraphics[width=1\textwidth]{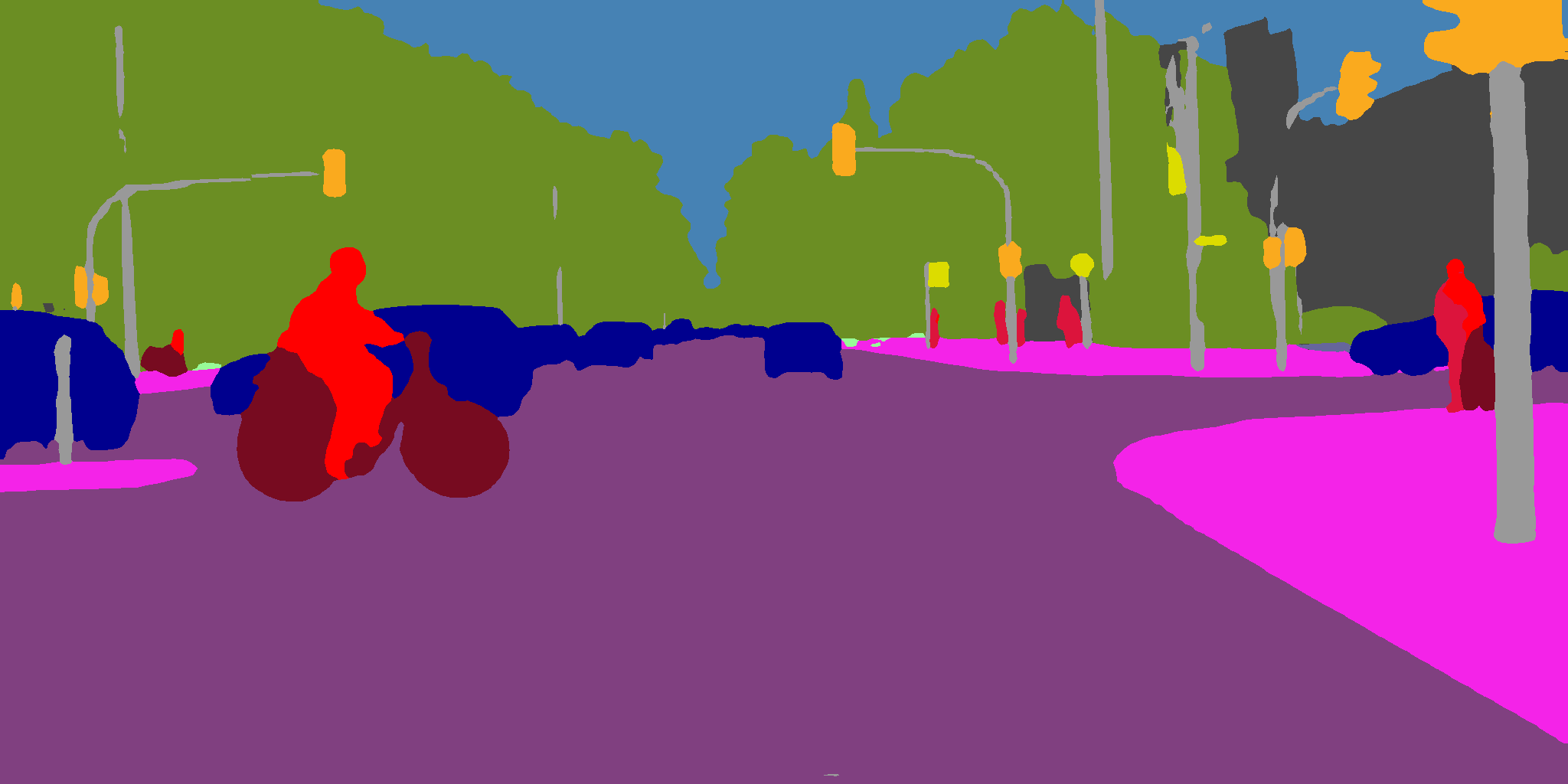}\\
\vspace{-2mm}
\includegraphics[width=1\textwidth]{munster_000000_000019_gtFine_color.png}\\
\end{minipage}
}
\hspace{-2mm}
\subfigure[]{
\begin{minipage}{0.23\linewidth}
\includegraphics[width=1\textwidth]{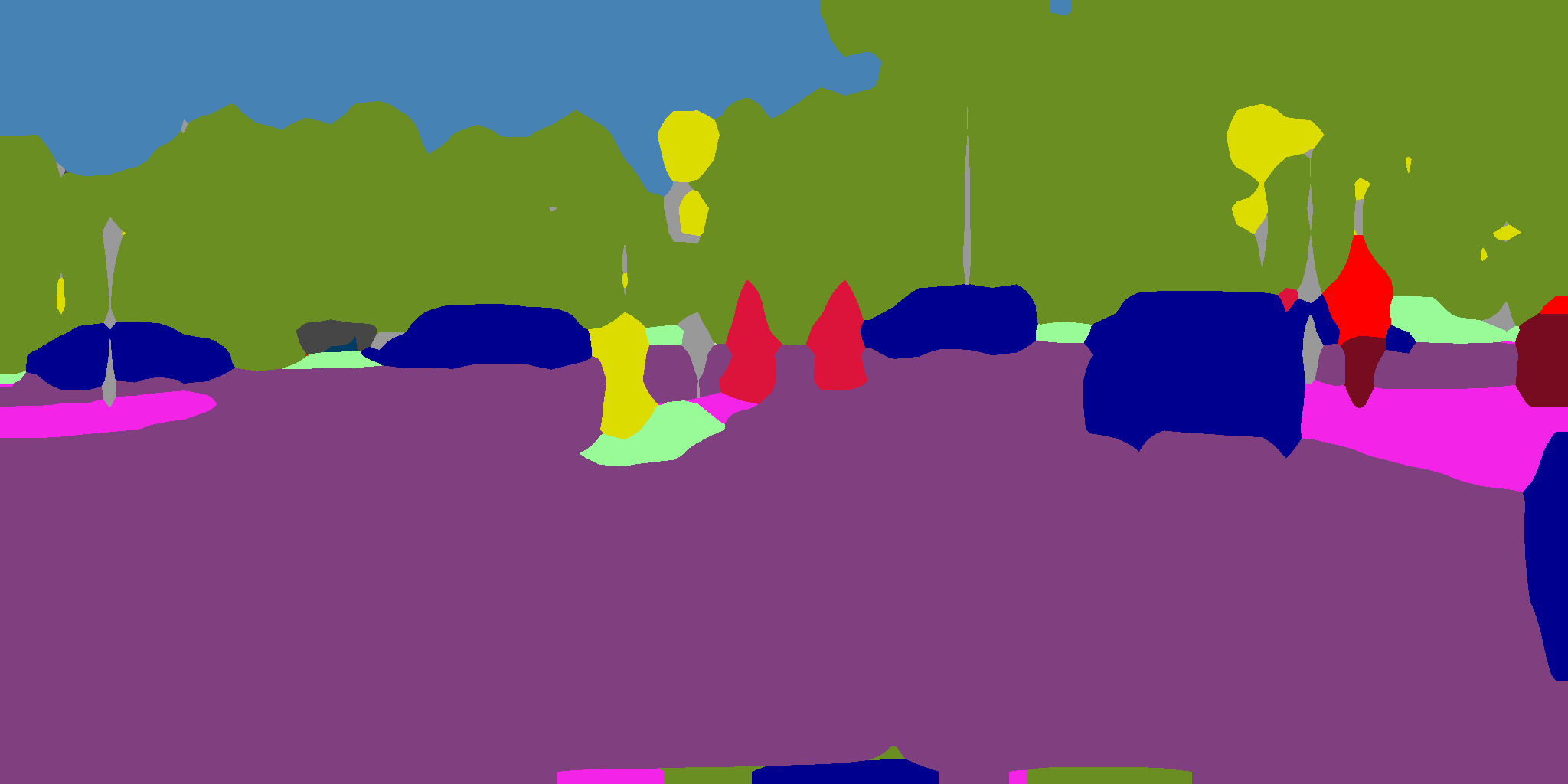}\\
\vspace{-2mm}
\includegraphics[width=1\textwidth]{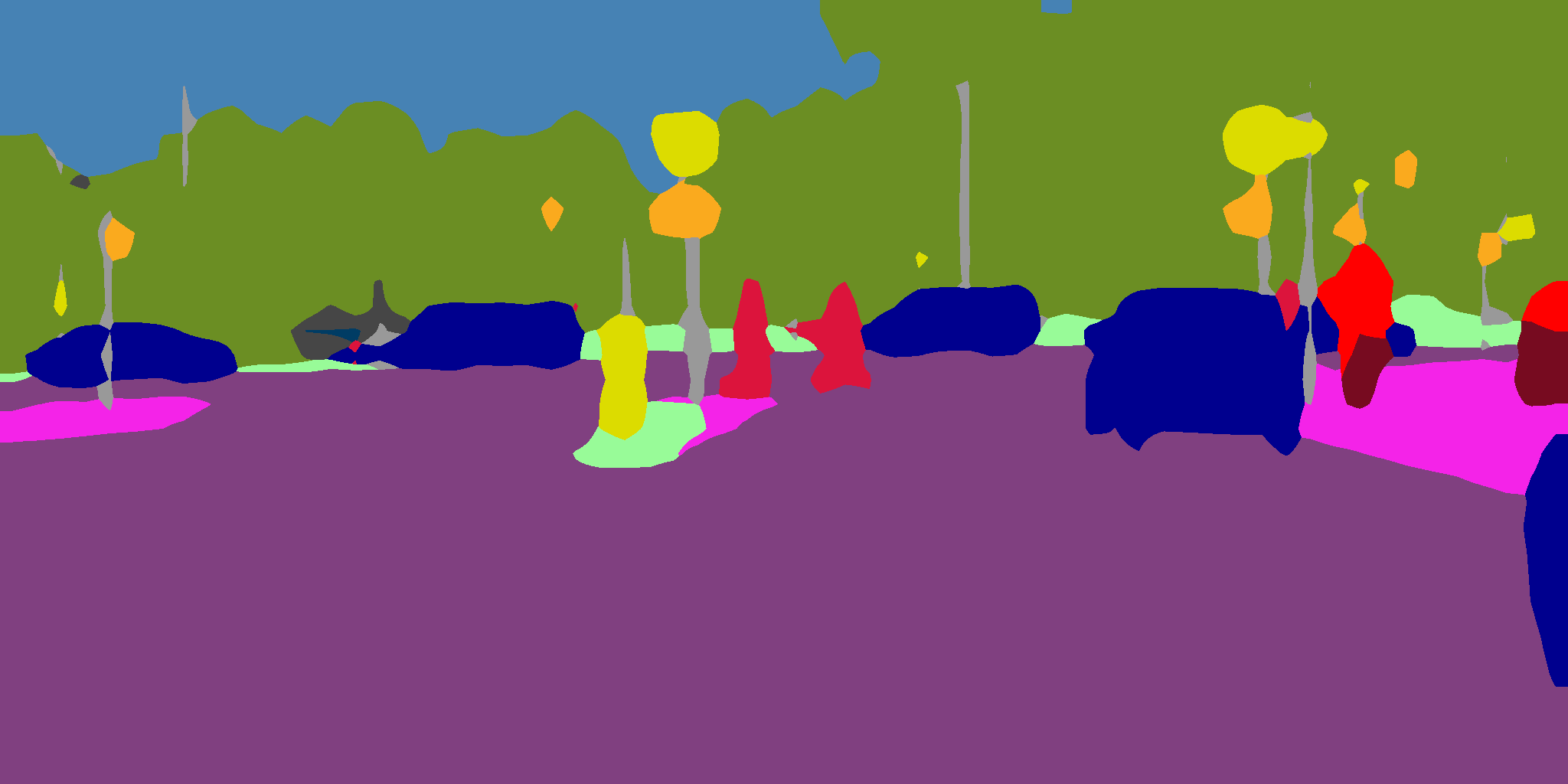}\\
\vspace{-2mm}
\includegraphics[width=1\textwidth]{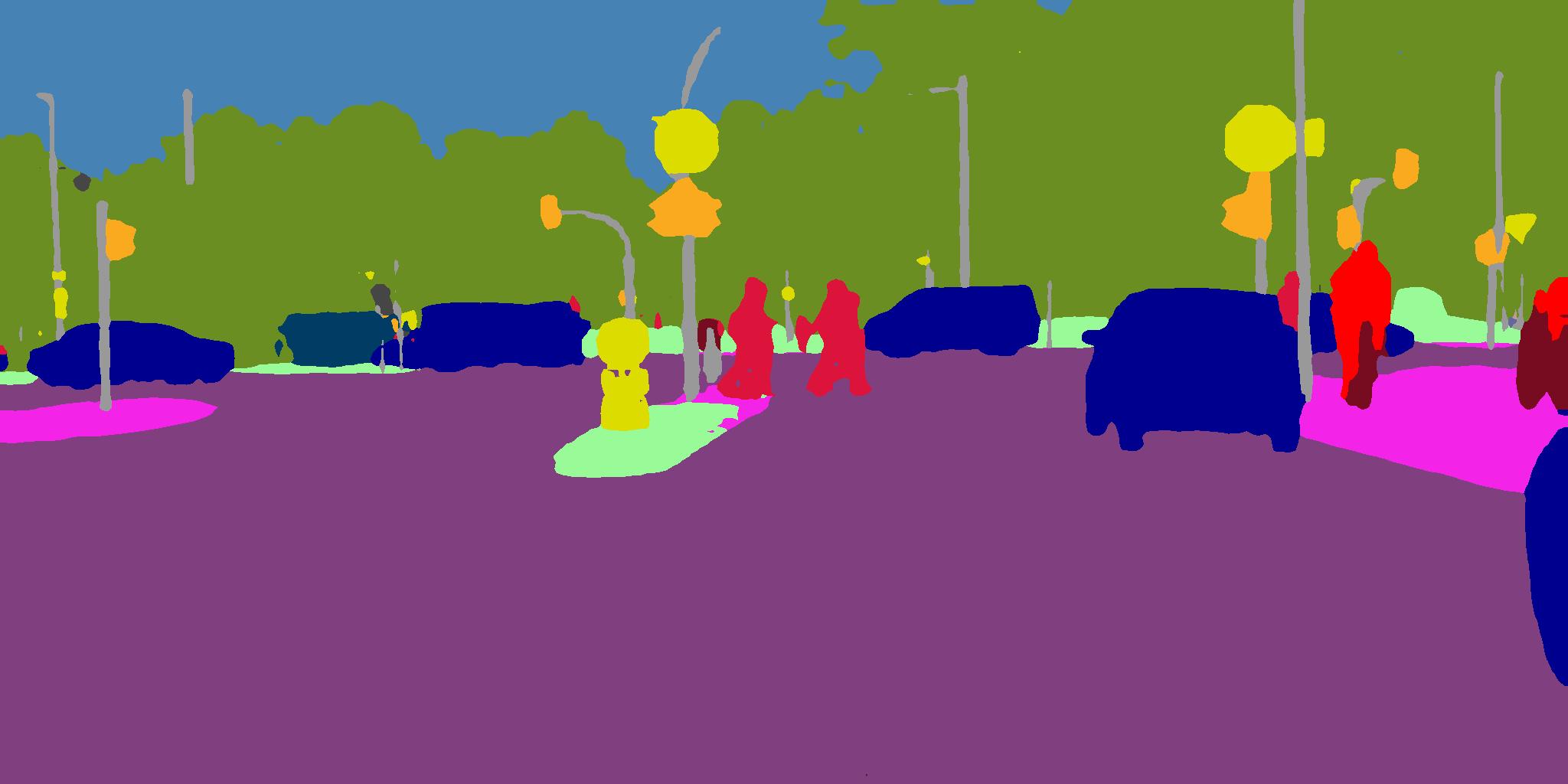}\\
\vspace{-2mm}
\includegraphics[width=1\textwidth]{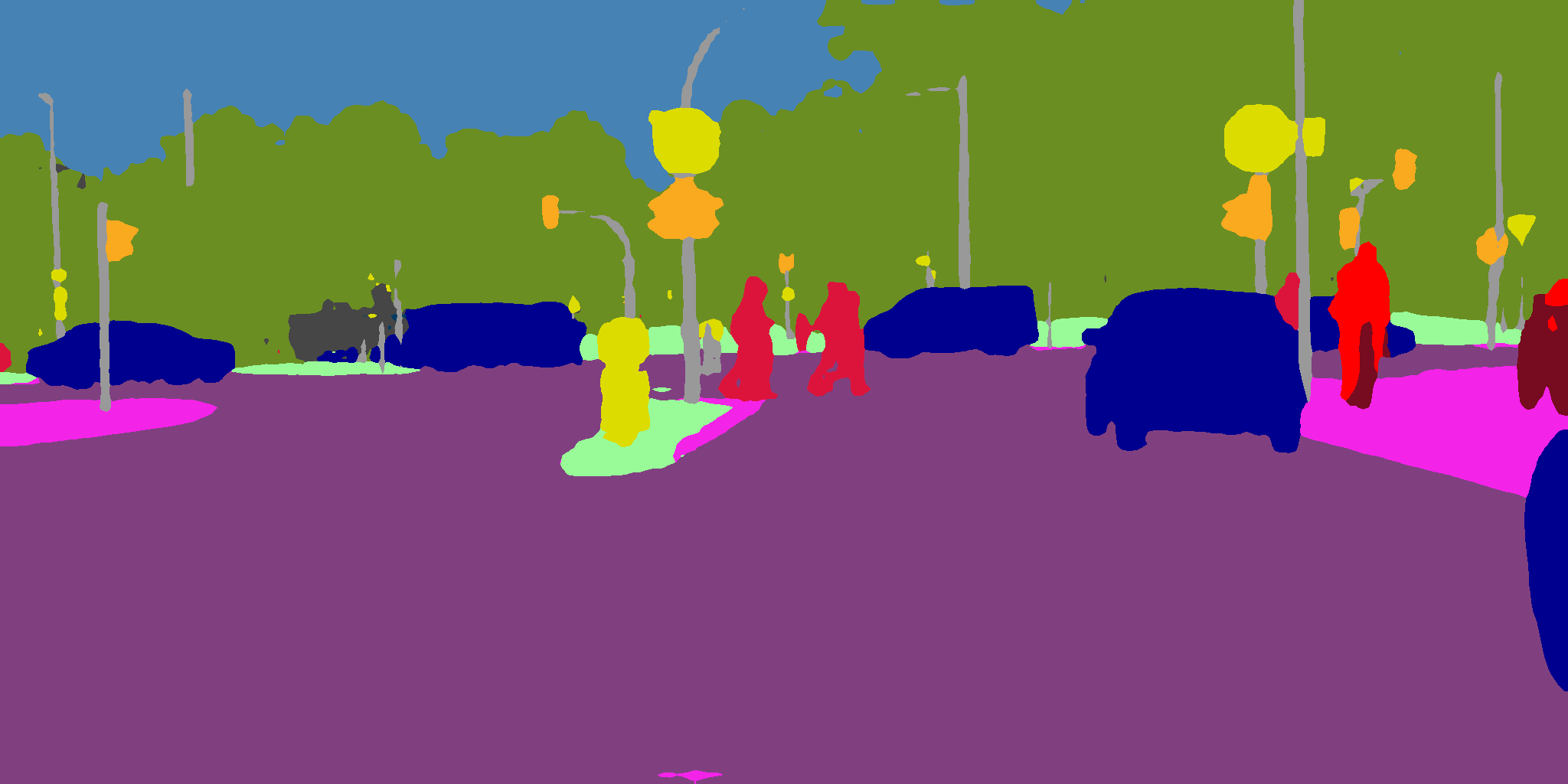}\\
\vspace{-2mm}
\includegraphics[width=1\textwidth]{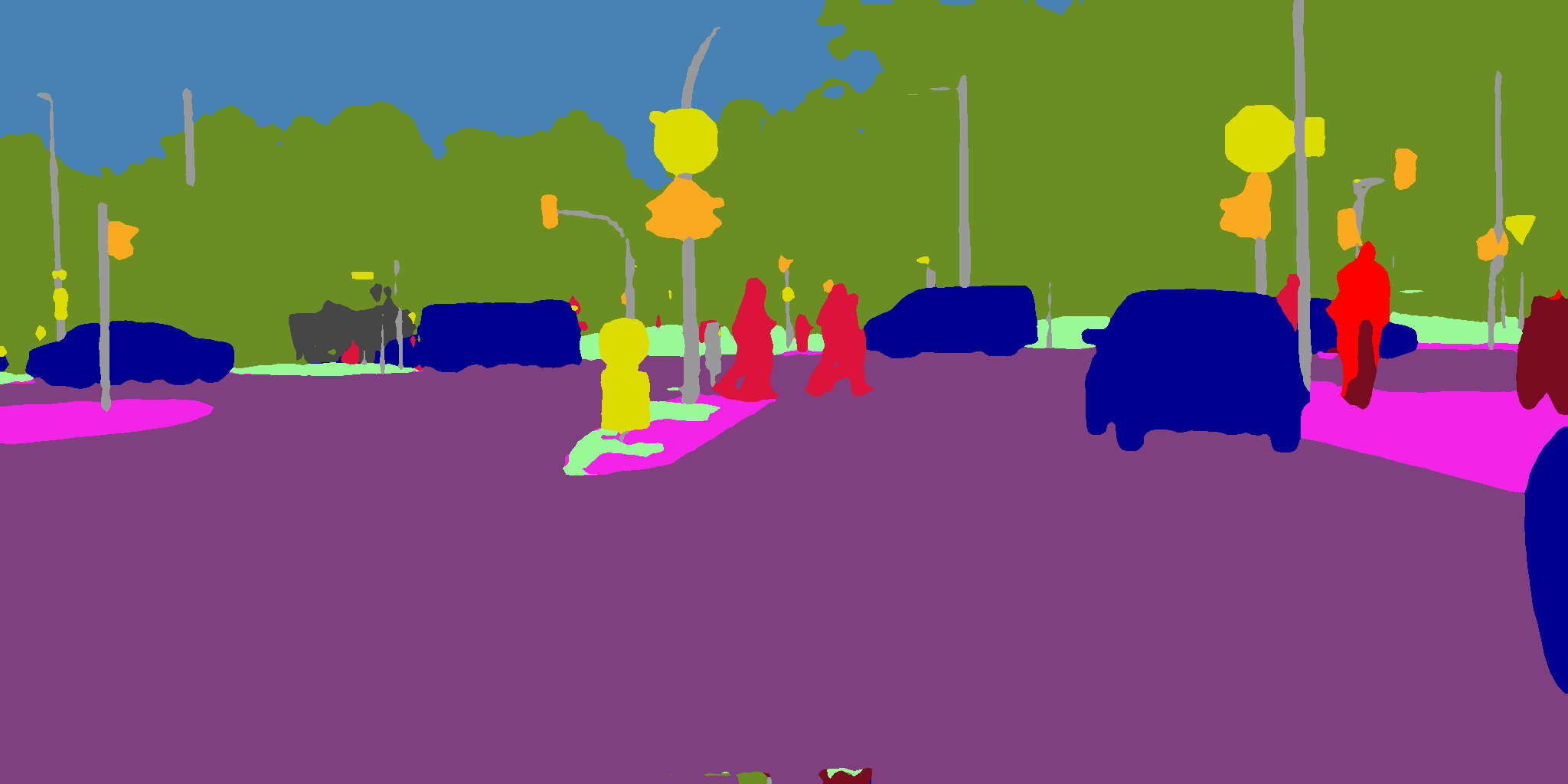}\\
\vspace{-2mm}
\includegraphics[width=1\textwidth]{munster_000166_000019_gtFine_color.png}\\
\end{minipage}
}
\hspace{-2mm}
\subfigure[]{
\begin{minipage}{0.23\linewidth}
\includegraphics[width=1\textwidth]{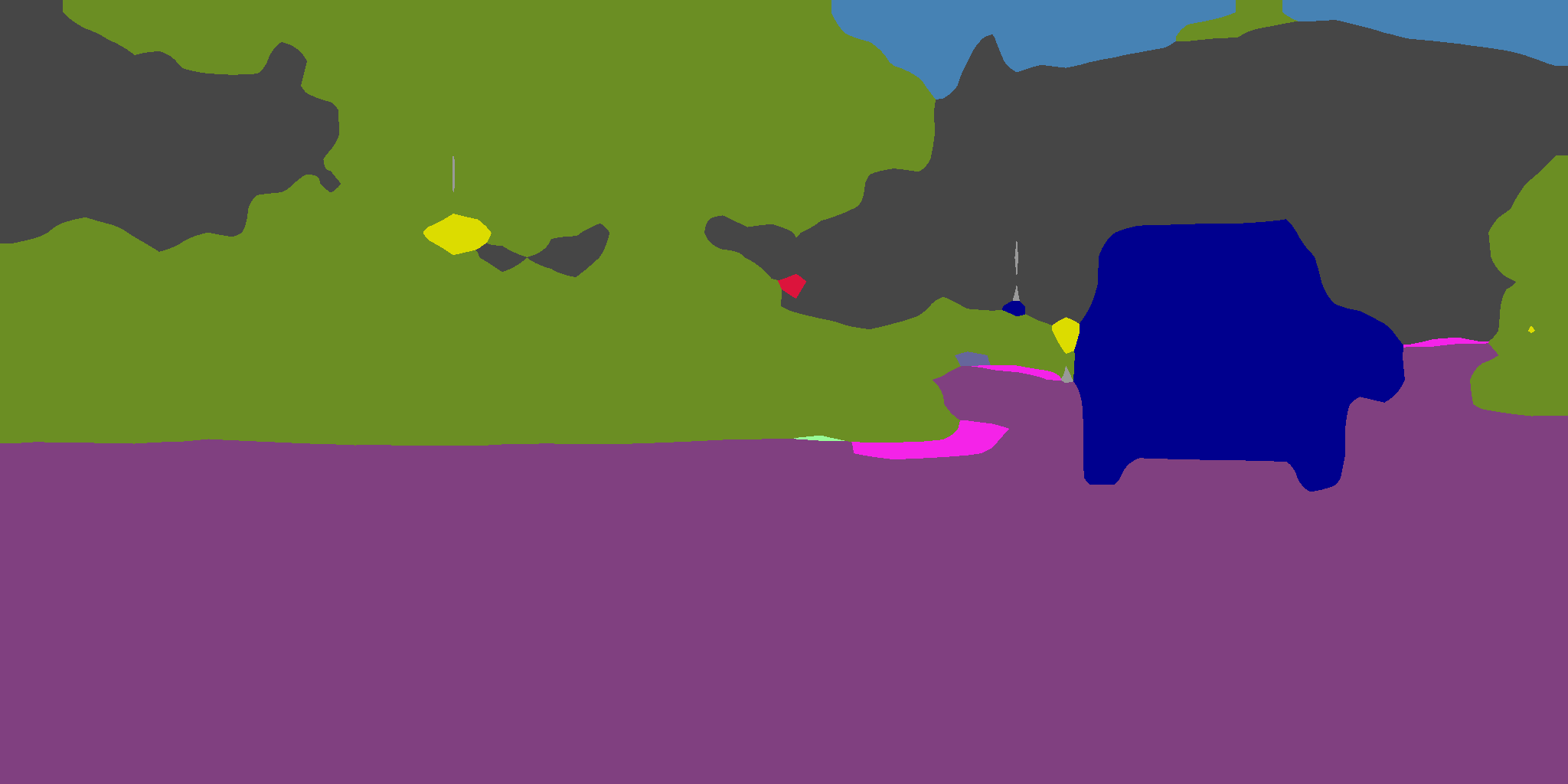}\\
\vspace{-2mm}
\includegraphics[width=1\textwidth]{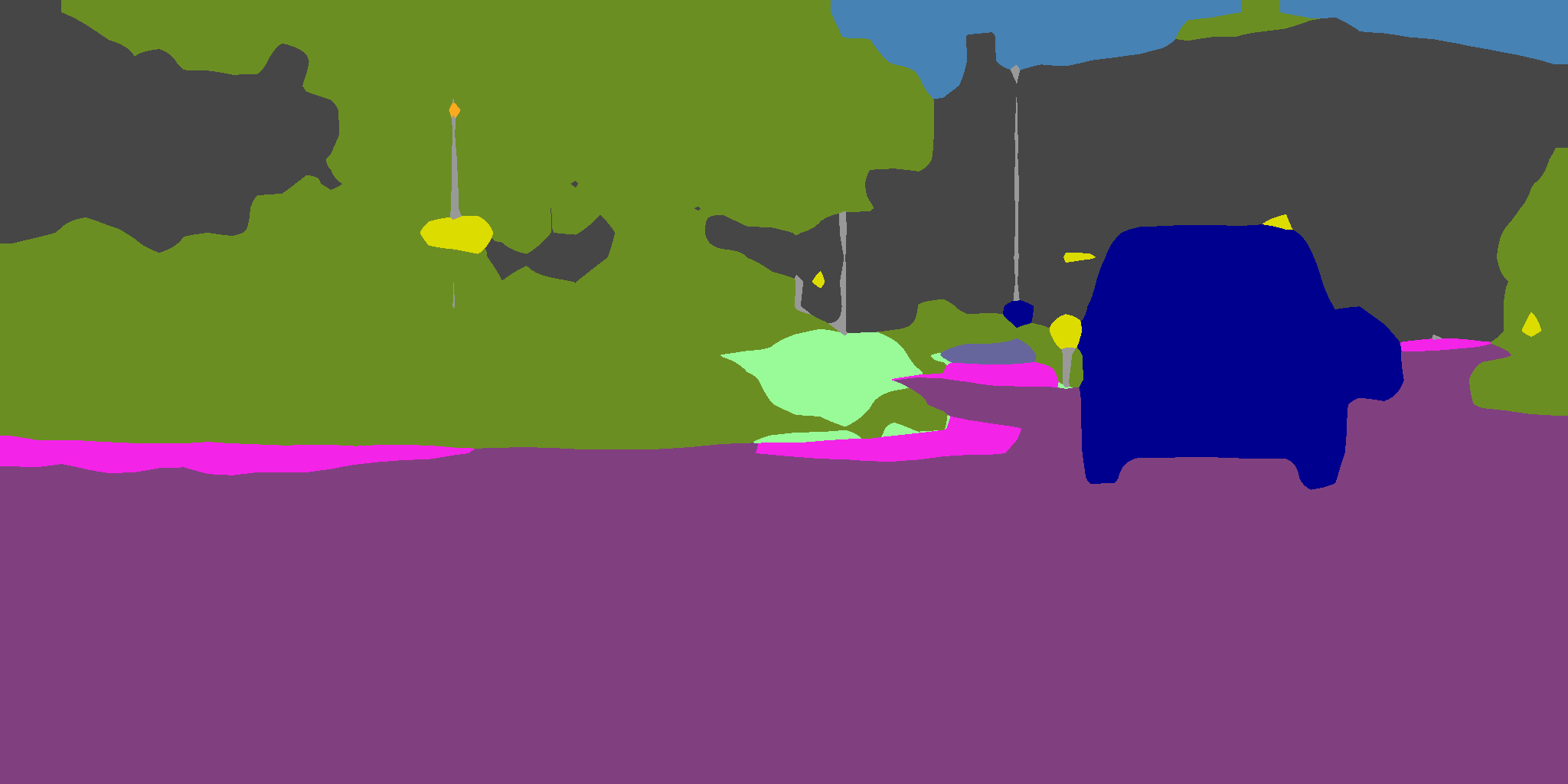}\\
\vspace{-2mm}
\includegraphics[width=1\textwidth]{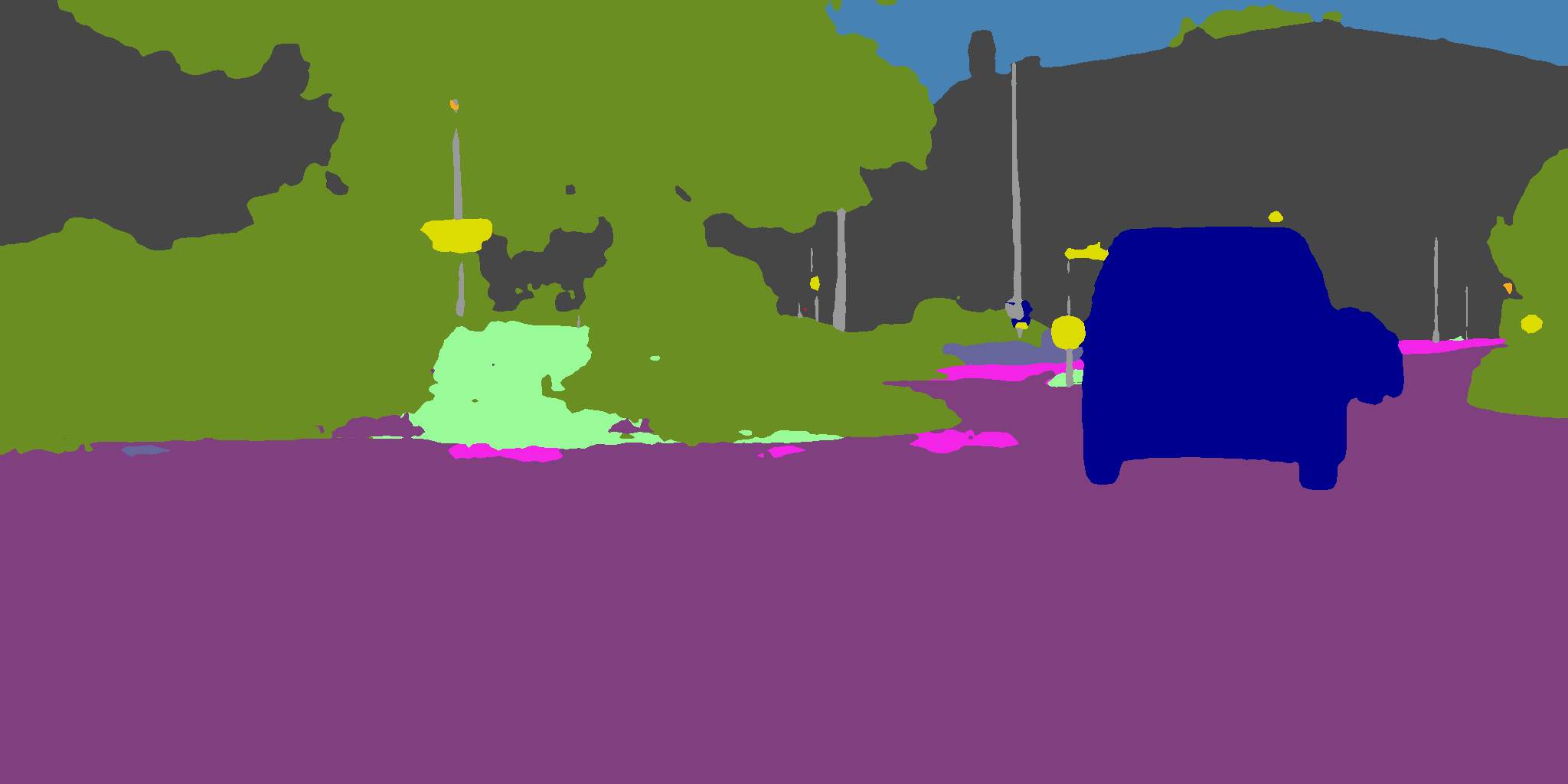}\\
\vspace{-2mm}
\includegraphics[width=1\textwidth]{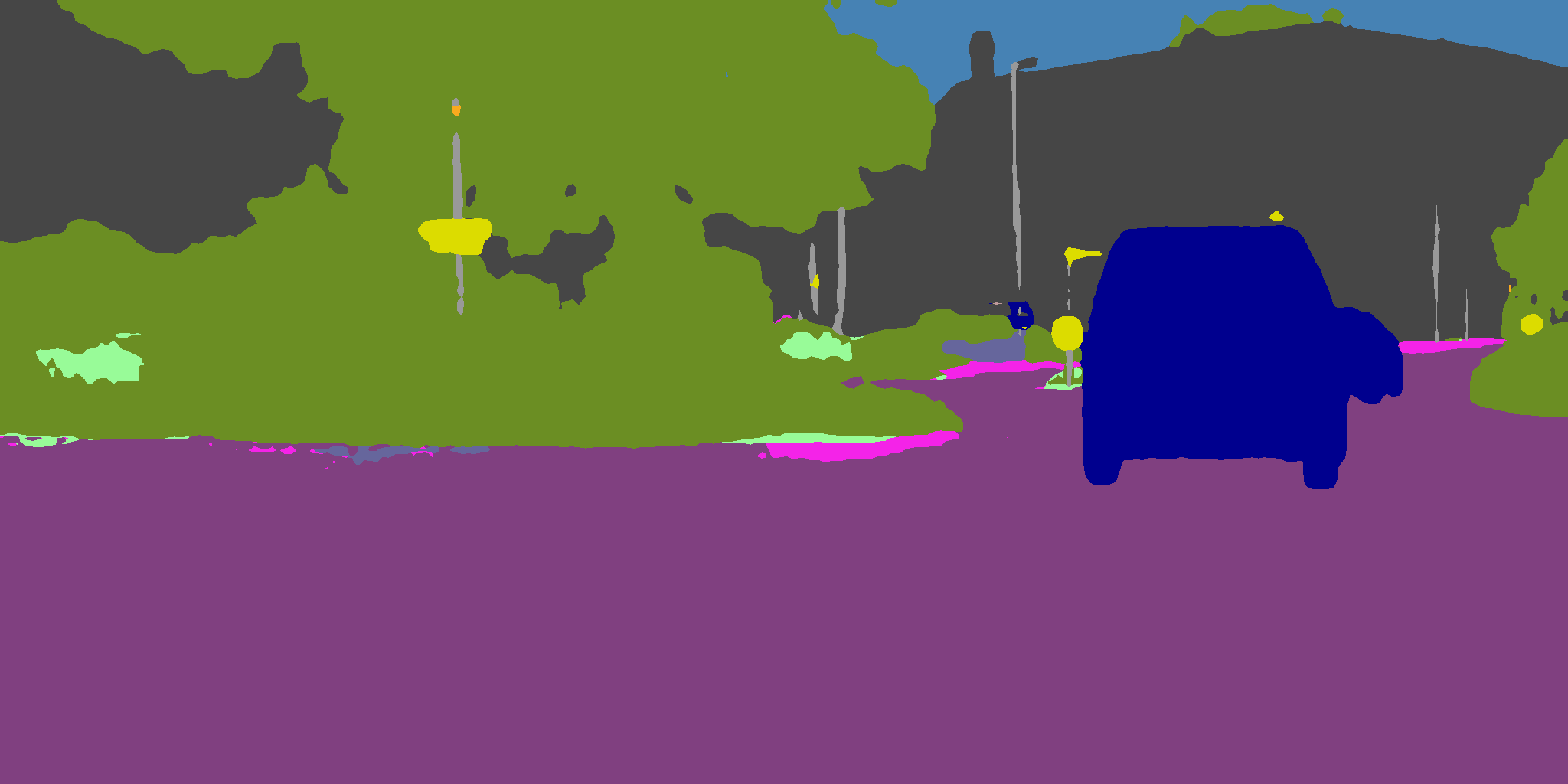}\\
\vspace{-2mm}
\includegraphics[width=1\textwidth]{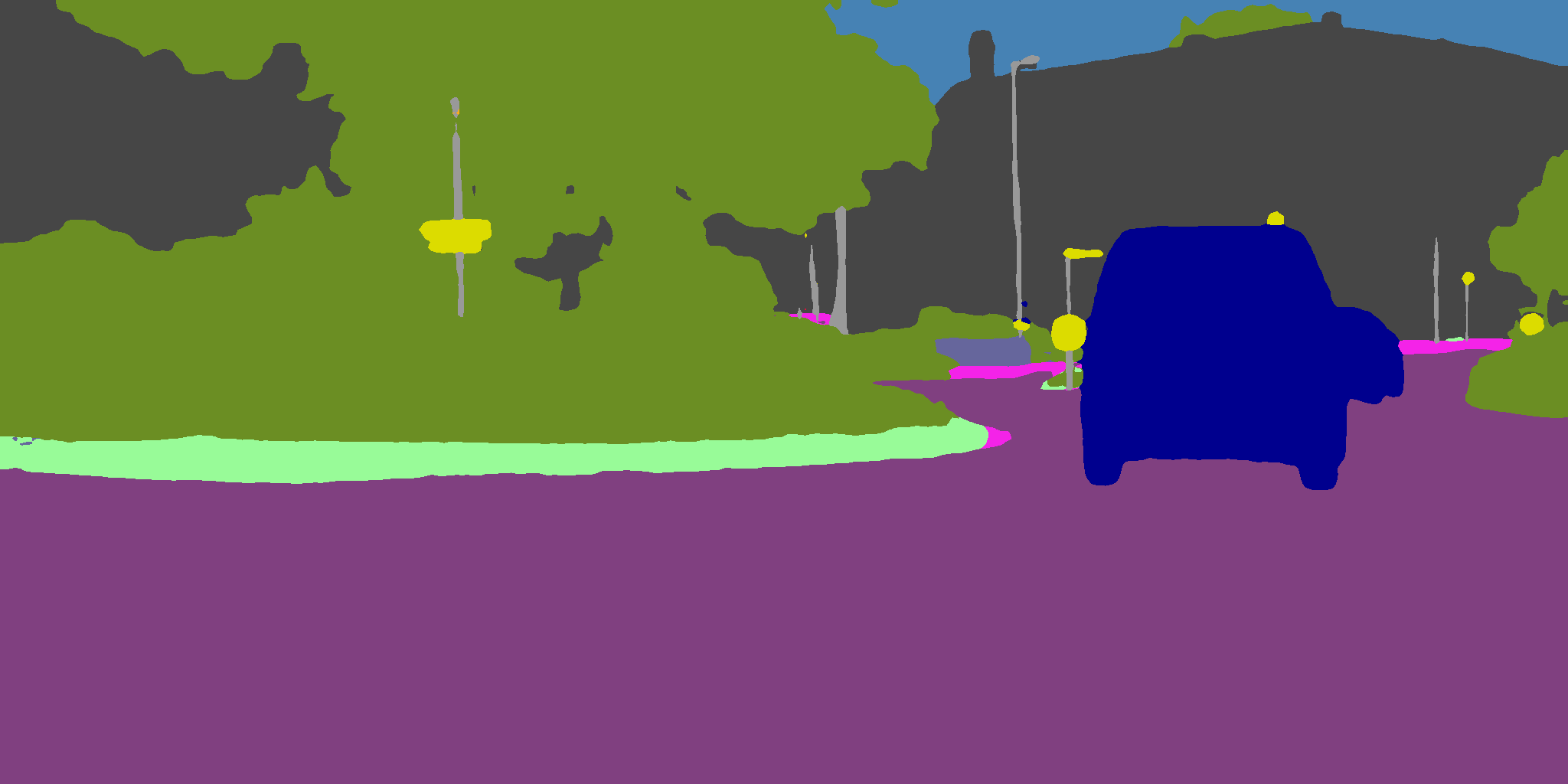}\\
\vspace{-2mm}
\includegraphics[width=1\textwidth]{lindau_000008_000019_gtFine_color.png}\\
\end{minipage}
}

\caption{Visual qualitative comparison results on Cityscapes validation set. Six lines from top to bottom represent baseline, STDC1+SE-ASPP, STDC1+CCBAM, STDC1+SE-ASPP+CCBAM, STDC1+SE-ASPP+CCBAM+Aux-loss and ground truth in turn. We can observe that our proposed model can learn more small target information.}
\label{fig8_ablation-study}
\end{figure*}


\subsection{Compare with State-of-the-arts}
In this section, we conduct experiments on the Cityscapes dataset and the CamVid dataset to demonstrate that the proposed model achieves superior results. Then, the algorithm is compared with other existing state-of-the-art algorithms on two benchmarks, Cityscapes and CamVid.
\paragraph{Results on Cityscapes.}
As listed in Table \ref{comparison with state of the art}, model name, input image resolution, backbone, mIoU and FPS of previous approaches are provided. Cross-CBAM-L1,Cross-CBAM-L2, respectively represent the input image resolution is $512\times1024$, $768\times1536$. Experimental results show that Cross-CBAM network achieve higher mIoU with excellent inference speed. Specifically, Cross-CBAM-M1 achieves $74.18\%$ mIoU and 240.9 FPS on the validation set, about two percentage points higher than the STDC network\cite{fan2021rethinking}, while retaining a competitive inference speed. Cross-CBAM-L1 achieves $76.04\%$ mIoU and $187.9$ FPS, which is a state-of-the-art trade-off between performance and speed. For a resolution of $768\times1536$, Cross-CBAM-L2 achieves the highest accuracy-$77.2\%$ mIoU and $88.6$FPS. Cross-CBAM-L2 achieves similar accuracy to LSPANet\cite{xiao2022real}, but the inference speed of Cross-CBAM-L2 is 31.9\% faster than LSPANet\cite{xiao2022real}. The Cross-CBAM series models achieves comparable accuracy to the PP-LiteSeg\cite{peng2022pp} series models while maintaining the superiority of inference speed. Furthermore, compared to SegNext\cite{guo2022segnext}, our model achieves 71.8\% faster speed than SegNext at the expense of only 0.8\% mIoU. Compared with SegFormer\cite{xie2021segformer}, although our model is slightly inferior to SegFormer\cite{xie2021segformer} in accuracy, it has a great improvement in speed. We increase inference speed by 82.8\%, from 25FPS to 88.6FPS. Therefore, it can be seen that the traditional convolution can also achieve results that are not inferior to the Transformer method, at least in the Cityscapes\cite{Cordts2016} dataset. Part of the segmentation results are shown in figure\ref{fig7_visual segmentation results}.

\paragraph{Results on CamVid.}
To further verify the generalization of the model, we conduct experiments on the CamVid dataset. Following the previous methods, the input resolution for training and testing is set to $960\times720$. As Table \ref{camvid-comparison with state of the art} shown, the model achieves $75.6\%$ mIoU and 146.8 FPS, higher than the STDC2-Seg\cite{fan2021rethinking}. When compared to light-weight networks (ENet, 
CAS, GAS, EDANet), Cross-CBAm stands out among them on both speed and accuracy. Although STDC and PP-LIteSeg are faster than Cross-CBAM in terms of inference speed, Cross-CBAM achieves higher accuracy.  Cross-CBAM-L is 2.6\% mIoU higher than LSPANet\cite{xiao2022real} and 4.1\% mIoU higher than MFNet\cite{ha2017mfnet}, which is a considerable improvement for real-time semantic segmentation. At the same time, we have also made a huge breakthrough in inference speed, and our proposed method is 49.9\% faster than LSPANet\cite{xiao2022real}. The comparison results demonstrate that Cross-CBAM has superior capability compared with other models.

\subsection{Ablation Study}
In this section, we conduct ablation experiments to validate the effectiveness of each component in Cross-CBAM. The Cross-CBAM-M1 model and the Cityscapes dataset are selected for the ablation study. Table \ref{ablation study} lists the results of the ablation experiments. The baseline model is STDC1 backbone+FCN segmentation head. The atrous rates of SE-ASPP in the ablation study are (1,3) and the number of channels of SE-ASPP is 256. Adding the SE-ASPP module improves mIoU by $4.48\%$. The CCBAM module in Cross-CBAM-M1 improves the mIoU by $3.89\%$. Simultaneous use of SE-ASPP module and CCBAM module increases mIoU by another $1.39\%$. Meanwhile, adding an auxiliary segmentation head slightly improves segmentation accuracy. Figure \ref{fig8_ablation-study} provides qualitative comparison results on the Cityscapes validation set. As modules are added one by one, the Cross-CBAM network can learn more detail information and becomes closer to the ground truth. 
\paragraph{Atrous Rates Ablation Experiments}
As we all know, the more the number of atrous convolutions, larger the atrous rates, and the larger captured receptive fields. But more atrous convolutions and larger atrous rates are not suitable for real-time semantic segmentation. In order to balance accuracy and inference speed, we conduct ablation experiments with dilation rates (1,3), (2,4), (3,5), (1,3,5). Table \ref{atrous rates} shows the results compared to different atrous rates. When the atrous rates are set to (1, 3), we achieve $ 74.19\%$ mIoU, 241.1FPS, 11.31G Flops and 12.21M parameters at a resolution of $512\times1024$. Atrous rates (3,5) achieve the highest mIoU, however, it corresponds to larger model, more parameters, and slower inference speed. Therefore, to balance accuracy and inference speed, the atrous rates of our model is set to (1,3).
\paragraph{Different SE-ASPP Channels}
In our intuitive impression, more channels in a neural network preserve more details, but in SE-ASPP module, the situation is different. As shown in Table \ref{se-aspp channels}, we initially set the channel of atrous convolution in SE-ASPP to 512, resulting in a mIoU of $73.86\%$. However, when the channels are set to 512, we achieve $74.18\%$ mIoU. Reducing the number of channels in SE-ASPP yields higher accuracy, meanwhile, reducing the number of channels also reduces the number of model parameters and improves inference speed. One possible reason is that more channels lead to redundancy and wrong combination of features. So when designing the model, we can change the way of thinking. It is not that the more channels the better.

\section{Conclusions}
In this paper, we choose the excellent STDC\cite{fan2021rethinking} network as the backbone of Cross-CBAM network. The STDC has a single path and adopts an aggregation of feature maps for image representation. We propose a new module for obtaining variable receptive fields at low computational cost, called Squeeze-and-Excitation Atrous Spatial Pyramid Pooling Module(SE-ASPP). Meanwhile, it can obtain multi-scale information with fewer parameters and faster inference speed. Furthermore, to strengthen feature representation efficiently, we present a contextual attention fusion module named CCBAM, which leverages high-level semantic features to guide low-level detail features. And the low-level features are complementary to the high-level features. Based on these innovative modules, extensive experiments are conducted to demonstrate the effectiveness of the proposed modules. Meanwhile, our real-time semantic segmentation network achieves state-of-the-art trade-off between accuracy and inference speed.

\printcredits

\section*{Acknowledgement}
This work is supported by the National Natural Science Foundation of China(61973120, 61673175, 61603139).
\bibliographystyle{cas-model2-names}

\bibliography{cas-refs}


\end{document}